\documentclass[10pt,twocolumn,letterpaper]{article}
\usepackage[pagenumbers]{cvpr}
\usepackage[T1]{fontenc}
\usepackage{fix-cm,textcomp}
\usepackage{booktabs,array,multirow,xspace,float}
\usepackage{amsmath,amssymb}
\usepackage{algorithm}
\usepackage[noend]{algpseudocode}
\algrenewcommand\alglinenumber[1]{\footnotesize #1}
\algrenewcommand\algorithmicthen{}
\usepackage{stfloats}
\usepackage{tikz,pgfplots}
\usetikzlibrary{arrows.meta,positioning,calc,fit,backgrounds,shapes.geometric,patterns,patterns.meta}
\pgfplotsset{compat=1.18}
\definecolor{gcG}{HTML}{356C99}      
\definecolor{gcR}{HTML}{A35F33}      
\definecolor{gcF}{HTML}{79597D}      
\colorlet{gcGk}{gcG}
\colorlet{gcRk}{gcR}
\colorlet{gcFk}{gcF}
\colorlet{gcGt}{gcG!30!white}        
\colorlet{gcRt}{gcR!30!white}
\colorlet{gcFt}{gcF!30!white}
\colorlet{gcGm}{gcG!58!white}        
\colorlet{gcRm}{gcR!58!white}
\colorlet{gcFm}{gcF!58!white}
\colorlet{gcOurs}{gcG!90!black}      
\definecolor{gcTarget}{HTML}{39CA60} 
\colorlet{gcGl}{gcGm}\colorlet{gcRl}{gcRm}\colorlet{gcFl}{gcFm}
\definecolor{gcExt}{RGB}{190,194,200}   
\definecolor{gcExtD}{RGB}{126,131,138}  
\definecolor{gcTxt}{RGB}{30,30,30}
\definecolor{gcSub}{RGB}{92,92,92}
\definecolor{gcGrid}{RGB}{228,229,231}

\newcommand{\fTitle}{\small\bfseries}
\newcommand{\fSub}{\scriptsize}
\newcommand{\fTick}{\scriptsize}
\newcommand{\fVal}{\footnotesize}
\newcommand{\fOurs}{\footnotesize\bfseries}
\newcommand{\fAxis}{\scriptsize}
\newcommand{\fLeg}{\footnotesize}

\tikzset{
  gc hatch/.style={pattern={Lines[angle=45,distance=2.1pt,line width=0.5pt]}, pattern color=#1},
  gc hatch/.default=gcExtD,
}
\pgfplotsset{
  gc axis/.style={
    scale only axis, clip=false,
    axis x line*=bottom, axis y line*=left,
    axis line style={gcExtD, line width=0.45pt},
    tick style={gcExtD, line width=0.4pt}, major tick length=2pt,
    xticklabel style={font=\fAxis, text=gcSub, inner sep=1.2pt},
    yticklabel style={font=\fAxis, text=gcSub, inner sep=1pt},
    label style={font=\fAxis, text=gcSub, inner sep=1pt},
    grid style={gcGrid, line width=0.4pt},
  },
}
\newcommand{\gcwidth}[1]{\path let \p1=(current bounding box.south west) in (\x1,\y1) -- ++(#1,0);}
\usepackage[pagebackref,breaklinks,colorlinks,citecolor=blue,linkcolor=blue,urlcolor=blue]{hyperref}
\makeatletter
\expandafter\def\csname gcv@sspi8@ctx\endcsname{73.69}
\expandafter\def\csname gcv@sspi8@zoom\endcsname{82.67}
\expandafter\def\csname gcv@sspi8@gc\endcsname{85.33}
\expandafter\def\csname gcv@sspi8@gcN\endcsname{1,349}
\expandafter\def\csname gcv@sspi8@n\endcsname{1,581}
\expandafter\def\csname gcv@sspi8@calls\endcsname{2.56}
\expandafter\def\csname gcv@sspi8@g_calls\endcsname{2.06}
\expandafter\def\csname gcv@sspi8@dzoomN\endcsname{42}
\expandafter\def\csname gcv@sspi8@cizoom\endcsname{[$+$1.58,\,$+$3.80]}
\expandafter\def\csname gcv@sspi8@exitsG\endcsname{1,191}
\expandafter\def\csname gcv@sspi8@errStable\endcsname{9.8}
\expandafter\def\csname gcv@sspi8@errEsc\endcsname{40.3}
\expandafter\def\csname gcv@sspi8@ao\endcsname{83.74}
\expandafter\def\csname gcv@sspi8@fullSaving\endcsname{36.11}
\expandafter\def\csname gcv@sspk8@gc\endcsname{85.01}
\expandafter\def\csname gcv@sspk8@calls\endcsname{2.55}
\expandafter\def\csname gcv@sspk8@g_calls\endcsname{2.06}
\expandafter\def\csname gcv@sspi8@fRep\endcsname{27}
\expandafter\def\csname gcv@sspi8@fReg\endcsname{9}
\expandafter\def\csname gcv@sspi8@fNet\endcsname{18}
\expandafter\def\csname gcv@sspi8@fnF\endcsname{99}
\expandafter\def\csname gcv@mvp@acc\endcsname{82.67}
\expandafter\def\csname gcv@mvp@calls\endcsname{4.00}
\expandafter\def\csname gcv@mvp@diffp\endcsname{2.66}
\expandafter\def\csname gcv@mvp@ci\endcsname{[$+$1.08,\,$+$4.17]}
\expandafter\def\csname gcv@sig@allscreenlinf\endcsname{0.718}
\expandafter\def\csname gcv@sig@allcrossmodel\endcsname{0.807}
\expandafter\def\csname gcv@sig@sspEscSelfAUROC\endcsname{0.655}
\expandafter\def\csname gcv@sig@sspEscCrossAUROC\endcsname{0.762}
\expandafter\def\csname gcv@sig@cixs\endcsname{[0.049,\,0.128]}
\expandafter\def\csname gcv@sig@secin\endcsname{72}
\expandafter\def\csname gcv@sig@secout\endcsname{43}
\expandafter\def\csname gcv@zr@sspi8breaks\endcsname{42}
\expandafter\def\csname gcv@zr@sspi8esc\endcsname{38}
\expandafter\def\csname gcv@zr@sspi8rec\endcsname{28}

\expandafter\def\csname gcv@erssp@contextExitCorrectAvailable\endcsname{1}
\expandafter\def\csname gcv@erssp@contextExitErrors\endcsname{4}
\expandafter\def\csname gcv@gatessp@acquired\endcsname{390}
\expandafter\def\csname gcv@gatessp@actualPp\endcsname{+1.52}
\expandafter\def\csname gcv@gatessp@uniformHighPp\endcsname{+0.89}
\expandafter\def\csname gcv@gatessp@uniformLowPp\endcsname{-0.32}
\expandafter\def\csname gcv@gatessp@uniformMeanPp\endcsname{+0.28}
\expandafter\def\csname gcv@sspi8@ccN\endcsname{17}
\expandafter\def\csname gcv@sspi8@ccNet\endcsname{1}
\expandafter\def\csname gcv@sspi8@selCalls\endcsname{2.49}
\expandafter\def\csname gcv@sspi8@stage2Gain\endcsname{24}

\expandafter\def\csname gcv@ord@i8acc\endcsname{83.93}
\expandafter\def\csname gcv@ord@i8calls\endcsname{2.54}
\expandafter\def\csname gcv@ord@i8large\endcsname{0.54}
\expandafter\def\csname gcv@ord@i8ciflag\endcsname{[$-$2.59,\,$-$0.13]}
\expandafter\def\csname gcv@ord@i8dflag\endcsname{22}
\expandafter\def\csname gcv@ord@i8saveflag\endcsname{74}
\expandafter\def\csname gcv@ord@i8downzoom\endcsname{62}
\expandafter\def\csname gcv@ord@k8acc\endcsname{83.68}
\expandafter\def\csname gcv@ord@k8calls\endcsname{2.56}
\expandafter\def\csname gcv@ord@k8large\endcsname{0.56}
\expandafter\def\csname gcv@ord@k8downzoom\endcsname{54}

\expandafter\def\csname gcv@erssp@gGateTP\endcsname{157}
\expandafter\def\csname gcv@erssp@gGateDeferred\endcsname{390}
\expandafter\def\csname gcv@erssp@gGateErrors\endcsname{274}
\expandafter\def\csname gcv@erssp@fRequests\endcsname{99}
\expandafter\def\csname gcv@erssp@fNovelCohort\endcsname{40}
\expandafter\def\csname gcv@erssp@fNovelCorrect\endcsname{6}
\expandafter\def\csname gcv@erssp@fRepairs\endcsname{27}
\expandafter\def\csname gcv@erssp@fRegressions\endcsname{9}
\expandafter\def\csname gcv@erssp@cascadeRegressions\endcsname{19}
\expandafter\def\csname gcv@erssp@gExitErrors\endcsname{117}
\expandafter\def\csname gcv@erssp@gExitNoneAvailable\endcsname{113}
\expandafter\def\csname gcv@erssp@agreeExitErrors\endcsname{60}
\expandafter\def\csname gcv@erssp@agreeExitCorrectAvailable\endcsname{26}
\expandafter\def\csname gcv@erssp@fExitErrors\endcsname{51}
\expandafter\def\csname gcv@erssp@fExitCorrectAvailable\endcsname{17}
\expandafter\def\csname gcv@erssp@totalErrors\endcsname{232}
\expandafter\def\csname gcv@erssp@totalCorrectAvailable\endcsname{48}
\expandafter\def\csname gcv@erssp@zoomRepairs\endcsname{184}
\expandafter\def\csname gcv@erssp@zoomRepairsKept\endcsname{165}

\expandafter\def\csname gcv@ivssp@FSelectedN\endcsname{99}
\expandafter\def\csname gcv@ivssp@FSelectedNet\endcsname{18}
\expandafter\def\csname gcv@ivssp@FSkippedN\endcsname{291}
\expandafter\def\csname gcv@ivssp@FSkippedNet\endcsname{-24}
\expandafter\def\csname gcv@ivssp@FUniformMeanOne\endcsname{-1.5}
\expandafter\def\csname gcv@ivssp@FUniformLow\endcsname{-10}
\expandafter\def\csname gcv@ivssp@FUniformHigh\endcsname{7}

\expandafter\def\csname gcv@sspi8@sel\endcsname{84.19}

\expandafter\def\csname gcv@mixosg@detail\endcsname{86.08}
\expandafter\def\csname gcv@mixosg@rdetail\endcsname{84.90}
\expandafter\def\csname gcv@mixosg@sel\endcsname{88.04}
\expandafter\def\csname gcv@mixosg@bestctx\endcsname{84.90}
\expandafter\def\csname gcv@mixosg@callsnof\endcsname{2.37}
\expandafter\def\csname gcv@mixosg@ao\endcsname{87.25}
\expandafter\def\csname gcv@mixosg@gc\endcsname{88.43}
\expandafter\def\csname gcv@mixosg@n\endcsname{510}
\expandafter\def\csname gcv@mixosg@correct\endcsname{451}
\expandafter\def\csname gcv@mixosg@calls\endcsname{2.42}
\expandafter\def\csname gcv@mixosg@saved\endcsname{39.28}
\expandafter\def\csname gcv@mixosg@stage3pp\endcsname{+0.39}
\expandafter\def\csname gcv@mixosg@rReq\endcsname{97}
\expandafter\def\csname gcv@mixosg@fReq\endcsname{21}
\expandafter\def\csname gcv@mixosg@fRep\endcsname{3}
\expandafter\def\csname gcv@mixosg@fReg\endcsname{1}
\expandafter\def\csname gcv@mixosg@ctxReq\endcsname{11}
\expandafter\def\csname gcv@mixosg@ctxNet\endcsname{0}
\expandafter\def\csname gcv@mixosg@rExitAvail\endcsname{15}
\expandafter\def\csname gcv@mixosg@gGateTP\endcsname{42}
\expandafter\def\csname gcv@mixosg@gGateErr\endcsname{71}
\expandafter\def\csname gcv@mixosg@selfAUROC\endcsname{0.707}
\expandafter\def\csname gcv@mixosg@crossAUROC\endcsname{0.796}
\expandafter\def\csname gcv@mixosg@budgetAcquired\endcsname{97}
\expandafter\def\csname gcv@mixosg@budgetActualPp\endcsname{+1.96}
\expandafter\def\csname gcv@mixosg@budgetUniformMeanPp\endcsname{+0.22}
\expandafter\def\csname gcv@mixosg@budgetUniformLowPp\endcsname{-0.39}
\expandafter\def\csname gcv@mixosg@budgetUniformHighPp\endcsname{+0.78}
\expandafter\def\csname gcv@mixosg@detailcilo\endcsname{0.95}
\expandafter\def\csname gcv@mixosg@detailcihi\endcsname{3.86}
\expandafter\def\csname gcv@mixuiv@ctx\endcsname{66.02}
\expandafter\def\csname gcv@mixuiv@detail\endcsname{69.72}
\expandafter\def\csname gcv@mixuiv@rdetail\endcsname{64.07}
\expandafter\def\csname gcv@mixuiv@sel\endcsname{72.28}
\expandafter\def\csname gcv@mixuiv@bestctx\endcsname{66.02}
\expandafter\def\csname gcv@mixuiv@callsnof\endcsname{2.62}
\expandafter\def\csname gcv@mixuiv@ao\endcsname{71.93}
\expandafter\def\csname gcv@mixuiv@gc\endcsname{72.20}
\expandafter\def\csname gcv@mixuiv@n\endcsname{5479}
\expandafter\def\csname gcv@mixuiv@correct\endcsname{3951}
\expandafter\def\csname gcv@mixuiv@calls\endcsname{2.73}
\expandafter\def\csname gcv@mixuiv@saved\endcsname{31.20}
\expandafter\def\csname gcv@mixuiv@pooled\endcsname{72.11}
\expandafter\def\csname gcv@mixuiv@stage2pp\endcsname{+2.56}
\expandafter\def\csname gcv@mixuiv@stage3pp\endcsname{-0.08}
\expandafter\def\csname gcv@mixuiv@rReq\endcsname{1728}
\expandafter\def\csname gcv@mixuiv@fReq\endcsname{595}
\expandafter\def\csname gcv@mixuiv@fRep\endcsname{70}
\expandafter\def\csname gcv@mixuiv@fReg\endcsname{75}
\expandafter\def\csname gcv@mixuiv@ctxReq\endcsname{150}
\expandafter\def\csname gcv@mixuiv@ctxRep\endcsname{20}
\expandafter\def\csname gcv@mixuiv@ctxReg\endcsname{10}
\expandafter\def\csname gcv@mixuiv@ctxNet\endcsname{10}
\expandafter\def\csname gcv@mixuiv@rExitAvail\endcsname{277}
\expandafter\def\csname gcv@mixuiv@fExitAvail\endcsname{173}
\expandafter\def\csname gcv@mixuiv@fBefore\endcsname{31.9}
\expandafter\def\csname gcv@mixuiv@fAfter\endcsname{31.1}
\expandafter\def\csname gcv@mixuiv@selfAUROC\endcsname{0.627}
\expandafter\def\csname gcv@mixuiv@crossAUROC\endcsname{0.692}
\expandafter\def\csname gcv@mixuiv@basic_fReq\endcsname{138}
\expandafter\def\csname gcv@mixuiv@basic_ctx\endcsname{71.16}
\expandafter\def\csname gcv@mixuiv@basic_detail\endcsname{72.63}
\expandafter\def\csname gcv@mixuiv@basic_sel\endcsname{75.45}
\expandafter\def\csname gcv@mixuiv@basic_full\endcsname{75.68}
\expandafter\def\csname gcv@mixuiv@functional_fReq\endcsname{149}
\expandafter\def\csname gcv@mixuiv@functional_ctx\endcsname{66.48}
\expandafter\def\csname gcv@mixuiv@functional_detail\endcsname{68.74}
\expandafter\def\csname gcv@mixuiv@functional_sel\endcsname{71.67}
\expandafter\def\csname gcv@mixuiv@functional_full\endcsname{71.78}
\expandafter\def\csname gcv@mixuiv@spatial_fReq\endcsname{308}
\expandafter\def\csname gcv@mixuiv@spatial_ctx\endcsname{60.41}
\expandafter\def\csname gcv@mixuiv@spatial_detail\endcsname{67.80}
\expandafter\def\csname gcv@mixuiv@spatial_sel\endcsname{69.72}
\expandafter\def\csname gcv@mixuiv@spatial_full\endcsname{69.15}
\expandafter\def\csname gcv@mixuiv@spatial_fNet\endcsname{-11}
\expandafter\def\csname gcv@mixuiv@budgetAcquired\endcsname{1728}
\expandafter\def\csname gcv@mixuiv@budgetActualPp\endcsname{+2.56}
\expandafter\def\csname gcv@mixuiv@budgetUniformMeanPp\endcsname{+0.69}
\expandafter\def\csname gcv@mixuiv@budgetUniformLowPp\endcsname{+0.33}
\expandafter\def\csname gcv@mixuiv@budgetUniformHighPp\endcsname{+1.04}
\expandafter\def\csname gcv@mixuiv@cilo\endcsname{-0.40}
\expandafter\def\csname gcv@mixuiv@cihi\endcsname{0.93}
\expandafter\def\csname gcv@mixuiv@detailcilo\endcsname{1.69}
\expandafter\def\csname gcv@mixuiv@detailcihi\endcsname{3.28}
\expandafter\def\csname gcv@mixuiv@selcilo\endcsname{1.82}
\expandafter\def\csname gcv@mixuiv@selcihi\endcsname{3.30}

\expandafter\def\csname gcv@venus@n\endcsname{5479}
\expandafter\def\csname gcv@venus@correct\endcsname{3200}
\expandafter\def\csname gcv@venus@pooled\endcsname{58.40}
\expandafter\def\csname gcv@venus@macro\endcsname{58.65}
\makeatother
\expandafter\def\csname gcv@erssp@fGateDeferred\endcsname{99}
\expandafter\def\csname gcv@erssp@fGateErrors\endcsname{133}
\expandafter\def\csname gcv@erssp@fGateTP\endcsname{69}
\expandafter\def\csname gcv@erssp@fIncomingAccuracy\endcsname{30.3}
\expandafter\def\csname gcv@erssp@fRegressionRate\endcsname{30.0}
\expandafter\def\csname gcv@erssp@fRepairRate\endcsname{39.1}
\expandafter\def\csname gcv@erssp@fReturnedAccuracy\endcsname{48.5}
\expandafter\def\csname gcv@erssp@gExitCorrectAvailable\endcsname{4}
\expandafter\def\csname gcv@erssp@rExitCorrectAvailable\endcsname{27}
\expandafter\def\csname gcv@erssp@rExitErrors\endcsname{64}
\expandafter\def\csname gcv@mixosg@fAfter\endcsname{52.4}
\expandafter\def\csname gcv@mixosg@fBefore\endcsname{42.9}
\expandafter\def\csname gcv@mixosg@fExitAvail\endcsname{3}
\expandafter\def\csname gcv@mixosg@fExitErrors\endcsname{10}
\expandafter\def\csname gcv@mixosg@fGateErr\endcsname{32}
\expandafter\def\csname gcv@mixosg@fGateTP\endcsname{12}
\expandafter\def\csname gcv@mixosg@fRegRate\endcsname{11.1}
\expandafter\def\csname gcv@mixosg@fRepairRate\endcsname{25.0}
\expandafter\def\csname gcv@mixosg@gExitAvail\endcsname{1}
\expandafter\def\csname gcv@mixosg@gExitErrors\endcsname{29}
\expandafter\def\csname gcv@mixosg@rExitErrors\endcsname{20}
\expandafter\def\csname gcv@mixosg@stage2pp\endcsname{+1.96}
\expandafter\def\csname gcv@mixosg@totalAvail\endcsname{19}
\expandafter\def\csname gcv@mixosg@totalErrors\endcsname{59}
\expandafter\def\csname gcv@mixuiv@fExitErrors\endcsname{410}
\expandafter\def\csname gcv@mixuiv@fGateErr\endcsname{899}
\expandafter\def\csname gcv@mixuiv@fGateTP\endcsname{405}
\expandafter\def\csname gcv@mixuiv@fRegRate\endcsname{39.5}
\expandafter\def\csname gcv@mixuiv@fRepairRate\endcsname{17.3}
\expandafter\def\csname gcv@mixuiv@gExitAvail\endcsname{7}
\expandafter\def\csname gcv@mixuiv@gExitErrors\endcsname{624}
\expandafter\def\csname gcv@mixuiv@gGateErr\endcsname{1662}
\expandafter\def\csname gcv@mixuiv@gGateTP\endcsname{1038}
\expandafter\def\csname gcv@mixuiv@rExitErrors\endcsname{494}
\expandafter\def\csname gcv@mixuiv@totalAvail\endcsname{457}
\expandafter\def\csname gcv@mixuiv@totalErrors\endcsname{1528}
\newcommand{\V}[2]{\ifcsname gcv@#1@#2\endcsname\csname gcv@#1@#2\endcsname\else\errmessage{Undefined verified value #1@#2}\fi}

\newcommand{\ErrorRecoverySuppTable}{%
\begin{table}[H]\centering\footnotesize\setlength{\tabcolsep}{2.5pt}
\begin{tabular*}{\columnwidth}{@{\extracolsep{\fill}}lrrr@{}}\toprule
Metric & SSP & OSG-R & UIV\\\midrule
$G\!\to\!R$ precision & \V{erssp}{gGateTP}/\V{erssp}{gGateDeferred} & \V{mixosg}{gGateTP}/\V{mixosg}{rReq} & \V{mixuiv}{gGateTP}/\V{mixuiv}{rReq}\\
$G\!\to\!R$ recall & \V{erssp}{gGateTP}/\V{erssp}{gGateErrors} & \V{mixosg}{gGateTP}/\V{mixosg}{gGateErr} & \V{mixuiv}{gGateTP}/\V{mixuiv}{gGateErr}\\
$R\!\to\!F$ precision & \V{erssp}{fGateTP}/\V{erssp}{fGateDeferred} & \V{mixosg}{fGateTP}/\V{mixosg}{fReq} & \V{mixuiv}{fGateTP}/\V{mixuiv}{fReq}\\
$R\!\to\!F$ recall & \V{erssp}{fGateTP}/\V{erssp}{fGateErrors} & \V{mixosg}{fGateTP}/\V{mixosg}{fGateErr} & \V{mixuiv}{fGateTP}/\V{mixuiv}{fGateErr}\\
\midrule
$F$ requests & \V{erssp}{fRequests} & \V{mixosg}{fReq} & \V{mixuiv}{fReq}\\
Before $F$: correct (\%) & \V{erssp}{fIncomingAccuracy} & \V{mixosg}{fBefore} & \V{mixuiv}{fBefore}\\
After $F$: correct (\%) & \V{erssp}{fReturnedAccuracy} & \V{mixosg}{fAfter} & \V{mixuiv}{fAfter}\\
$F$ repair rate (\%) & \V{erssp}{fRepairRate} & \V{mixosg}{fRepairRate} & \V{mixuiv}{fRepairRate}\\
$F$ regression rate (\%) & \V{erssp}{fRegressionRate} & \V{mixosg}{fRegRate} & \V{mixuiv}{fRegRate}\\
\bottomrule\end{tabular*}
\caption{\textbf{Gate detection and recovery.} SSP uses I32/I8/I32; OSG-R and UIV use Q36/Q38/Q38. First-gate precision counts standalone Two-pass errors among escalations, and recall counts escalated standalone Two-pass errors among all such errors. Finalizer recall counts escalated Stage~2 errors among all Stage~2 errors; repair and regression rates use incorrect and correct Stage~2 outputs as denominators.}
\label{tab:s_error_recovery}\end{table}}

\newcommand{\gc}{GroundCascade\xspace}
\newcommand{\smallhead}[1]{\noindent\textbf{#1}\ }
\DeclareMathOperator*{\argmin}{arg\,min}
\newcommand{\papertitle}{GroundCascade: Selective Second Opinions for GUI Grounding}
\def\confName{CVPR}
\def\confYear{2027}
\hypersetup{pdftitle={GroundCascade: Selective Second Opinions for GUI Grounding},pdfsubject={GUI grounding, selective inference, model cascades},pdfkeywords={GUI grounding, visual refinement, model cascades},pdfauthor={}}

\title{\papertitle}
\author{%
Adrian Phoulady\\
University of Connecticut\\
Storrs, CT, USA\\
{\small\texttt{adrian.phoulady@uconn.edu}}
\and
Hady Ahmady Phoulady\\
Embry-Riddle Aeronautical University\\
Daytona Beach, FL, USA\\
{\small\texttt{ahmadyph@erau.edu}}
}
\hypersetup{pdfauthor={Adrian Phoulady and Hady Ahmady Phoulady}}

\begin{document}
\newcommand{\GCTeaserCaption}{%
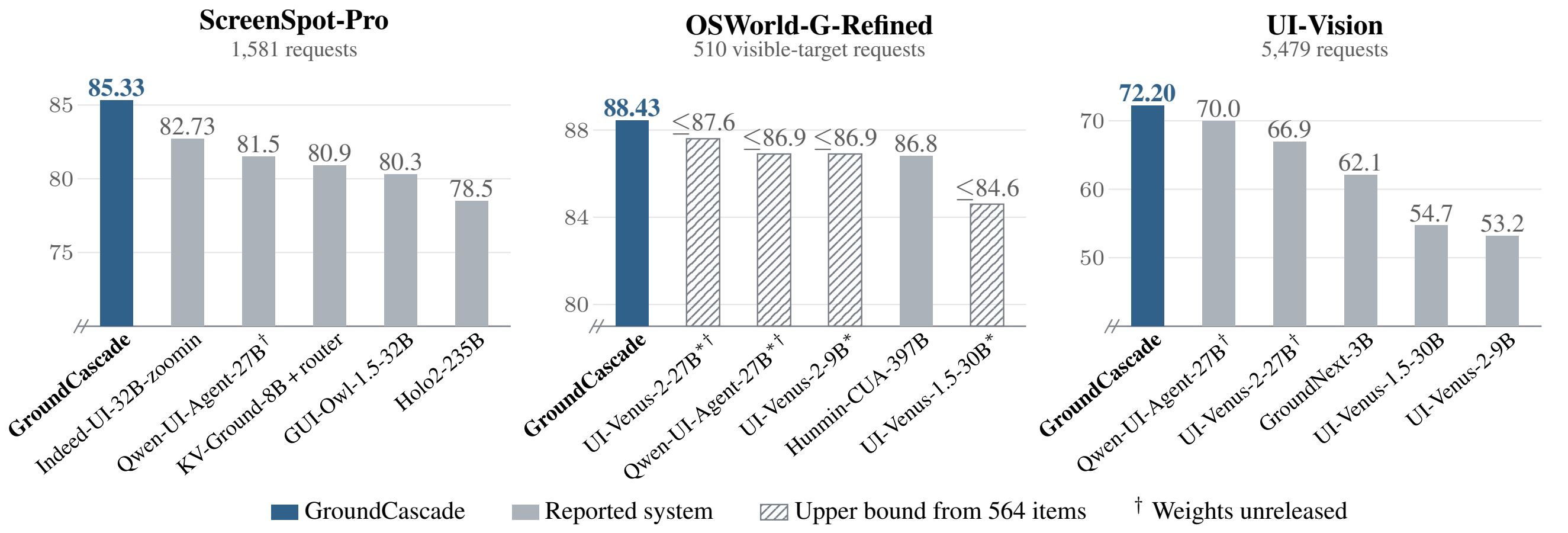
\textbf{GroundCascade versus reported GUI-focused systems} (accuracy, \%).
  I32/I8/I32 (max 32B) is used on SSP and Q36/Q38/Q38 (max 27B) elsewhere.
  $^*$ Hatched $\leq$ bars conditionally bound OSG-R accuracy if published scores are
  micro-accuracies over 564 requests; they are not measured 510-item scores.
  UIV is category-balanced; external protocols may differ~\cite{indeedui,qwenuiagent,zoomconsistency,guiowl15,holo2,venus2,hunmin_cua,venus15,groundnext}.}

\maketitle
\begin{figure*}[b]
  \centering
  \begingroup
\colorlet{tOurs}{gcOurs}            
\definecolor{tPub}{RGB}{172,178,186}
\definecolor{tPubD}{RGB}{120,126,134}

\pgfplotsset{
  tpanel/.style={
    scale only axis, width=4.84cm, height=2.80cm,
    xmin=0.45, xmax=6.55, xtick={1,...,6}, clip=false,
    axis x line*=bottom, axis y line*=left, y axis line style={draw=none},
    x axis line style={gcExtD, line width=0.5pt},
    xtick style={draw=none}, ytick style={draw=none},
    ymajorgrids, grid style={gcGrid, line width=0.4pt},
    yticklabel style={font=\fAxis, text=gcSub, inner sep=1pt},
    xticklabel style={font=\fTick, rotate=40, anchor=north east,
                      inner sep=0.8pt, yshift=-1.0pt, xshift=1.6pt},
  },
}
\newlength{\hw}\setlength{\hw}{0.185cm}   
\tikzset{
  pub/.style   ={fill=tPub, draw=none},
  bound/.style ={fill=white, draw=tPubD, line width=0.45pt,
                 pattern={Lines[angle=45,distance=2.0pt,line width=0.5pt]}, pattern color=tPubD},
  ours/.style  ={fill=tOurs, draw=none},
  val/.style   ={font=\fVal, text=gcSub, inner sep=1.1pt, anchor=south},
  oursval/.style={font=\fOurs, text=tOurs, inner sep=1.1pt, anchor=south},
}
\newcommand{\tbar}[5]{%
  \path[#3] ([xshift=-\hw]axis cs:#1,\ymn) rectangle ([xshift=\hw]axis cs:#1,#2);
  \node[#4] at (axis cs:#1,#2) {#5};}
\newcommand{\tours}[1]{\tbar{1}{#1}{ours}{oursval}{#1}}
\newcommand{\heading}[2]{%
  \node[anchor=south, inner sep=0pt, font=\fSub, text=gcSub] (sub) at ([yshift=4.5pt]rel axis cs:0.5,1) {#2};
  \node[anchor=south, inner sep=0pt, font=\fTitle] at ([yshift=1.6pt]sub.north) {#1};}
\newcommand{\axisbreak}{%
  \draw[gcExtD, line width=0.45pt] ([xshift=-1.2pt,yshift=-2.2pt]rel axis cs:0,0) -- ++(2.4pt,4.4pt)
                                   ([xshift=1.0pt,yshift=-2.2pt]rel axis cs:0,0) -- ++(2.4pt,4.4pt);}
\begin{tikzpicture}[x=1cm,y=1cm]

\def\ymn{70}
\begin{axis}[tpanel, name=ssp, at={(0,0)}, ymin=70, ymax=87.0, ytick={75,80,85},
  xticklabels={\textbf{GroundCascade}, Indeed-UI-32B-zoomin, Qwen-UI-Agent-27B$^\dagger$,
               KV-Ground-8B\,+\,router, GUI-Owl-1.5-32B, Holo2-235B}]
  \tours{85.33}
  \tbar{2}{82.73}{pub}{val}{82.73}
  \tbar{3}{81.5}{pub}{val}{81.5}
  \tbar{4}{80.9}{pub}{val}{80.9}
  \tbar{5}{80.3}{pub}{val}{80.3}
  \tbar{6}{78.5}{pub}{val}{78.5}
  \axisbreak
  \heading{ScreenSpot-Pro}{1,581 requests}
\end{axis}

\def\ymn{79}
\begin{axis}[tpanel, name=osg, at={($(ssp.south east)+(0.92cm,0)$)},
  ymin=79, ymax=90.5, ytick={80,84,88},
  xticklabels={\textbf{GroundCascade}, UI-Venus-2-27B$^{*\dagger}$,
               Qwen-UI-Agent-27B$^{*\dagger}$, UI-Venus-2-9B$^*$,
               Hunmin-CUA-397B, UI-Venus-1.5-30B$^*$}]
  \tours{88.43}
  \tbar{2}{87.6}{bound}{val}{$\leq$87.6}
  \tbar{3}{86.9}{bound}{val}{$\leq$86.9}
  \tbar{4}{86.9}{bound}{val}{$\leq$86.9}
  \tbar{5}{86.80}{pub}{val}{86.8}
  \tbar{6}{84.6}{bound}{val}{$\leq$84.6}
  \axisbreak
  \heading{OSWorld-G-Refined}{510 visible-target requests}
\end{axis}

\def\ymn{40}
\begin{axis}[tpanel, name=uiv, at={($(osg.south east)+(0.92cm,0)$)}, ymin=40, ymax=76.6, ytick={50,60,70},
  xticklabels={\textbf{GroundCascade}, Qwen-UI-Agent-27B$^\dagger$,
               UI-Venus-2-27B$^\dagger$, GroundNext-3B, UI-Venus-1.5-30B, UI-Venus-2-9B}]
  \tours{72.20}
  \tbar{2}{70.0}{pub}{val}{70.0}
  \tbar{3}{66.9}{pub}{val}{66.9}
  \tbar{4}{62.1}{pub}{val}{62.1}
  \tbar{5}{54.7}{pub}{val}{54.7}
  \tbar{6}{53.2}{pub}{val}{53.2}
  \axisbreak
  \heading{UI-Vision}{5,479 requests}
\end{axis}

\node[inner sep=0pt, anchor=north, font=\fLeg]
  at ($(ssp.south west)!0.5!(uiv.south east)+(0,-1.90cm)$) {%
  \def\sw{0.30cm}\def\sh{0.17cm}%
  \tikz[baseline=-0.6ex]\fill[tOurs] (0,-\sh/2) rectangle (\sw,\sh/2);~GroundCascade\hspace{0.52cm}%
  \tikz[baseline=-0.6ex]\fill[tPub] (0,-\sh/2) rectangle (\sw,\sh/2);~Reported system\hspace{0.52cm}%
  \tikz[baseline=-0.6ex]\path[bound] (0,-\sh/2) rectangle (\sw,\sh/2);~Upper bound from 564 items\hspace{0.52cm}%
  $^\dagger$ Weights unreleased};

\path let \p1=(current bounding box.south west), \p2=(current bounding box.north east) in \pgfextra{\typeout{BBOX: \x1, \x2, \y1, \y2}};
\gcwidth{6.875in}   
\end{tikzpicture}
\endgroup
  \caption{\GCTeaserCaption}
  \label{fig:teaser}
\end{figure*}

\begin{abstract}
Graphical user interface (GUI) grounding maps a screenshot and a natural-language instruction to the screen location where an agent should act. Zoomed views can reveal small targets but may lose the context needed to identify them. We introduce \gc, a training-free controller that uses agreement between a grounder's full-screen and zoomed predictions to decide when to consult a complementary model. It selects between zoomed predictions while preserving corroborated full-screen locations. An optional Finalizer predicts from an unmarked joint crop when disagreement remains (Full); omitting it yields the two-stage Selective variant. \mbox{GroundCascade} reaches \V{sspi8}{gc}\% on ScreenSpot-Pro with Indeed-UI-32B/8B/32B and \V{mixosg}{gc}\% on the visible-target OSWorld-G-Refined subset with Qwen3.6-27B/3.8-27B/3.8-27B, gains of 2.66 and 2.35 percentage points over the stronger standalone two-pass grounder. On UI-Vision, GroundCascade reaches \V{mixuiv}{gc}\% category-balanced accuracy, with Selective slightly higher at \V{mixuiv}{sel}\%. Across these three evaluations, \mbox{GroundCascade} in Full mode uses 2.42--2.73 calls per request, 31--39\% fewer than unconditionally acquiring both models' two-pass predictions. At matched second-model call budgets, selective acquisition gains 1.52--2.56 points over standalone Two-pass, compared with 0.22--0.69 points from stratified random acquisition. GroundCascade leads the official ScreenSpot-Pro leaderboard and exceeds reported GUI-specialist comparisons on OSWorld-G-Refined and UI-Vision, subject to differences in external evaluation protocols.
\end{abstract}

\section{Introduction}
\label{sec:intro}
GUI grounding maps a natural-language instruction and screenshot to where an agent should act. Professional interfaces contain small icons and similar labels~\cite{screenspot,uivision}. Two-pass grounders magnify a crop around an initial full-screen prediction to resolve fine details~\cite{zoomclick,maiui,kvground,indeedui}. Yet a crop can hide the toolbar or panel that identifies a target, spoiling a correct full-screen prediction~\cite{zoomclick,uizoomer}. A complementary model can repair errors~\cite{zoomconsistency}, but need not be better; using both two-pass models can take four calls.

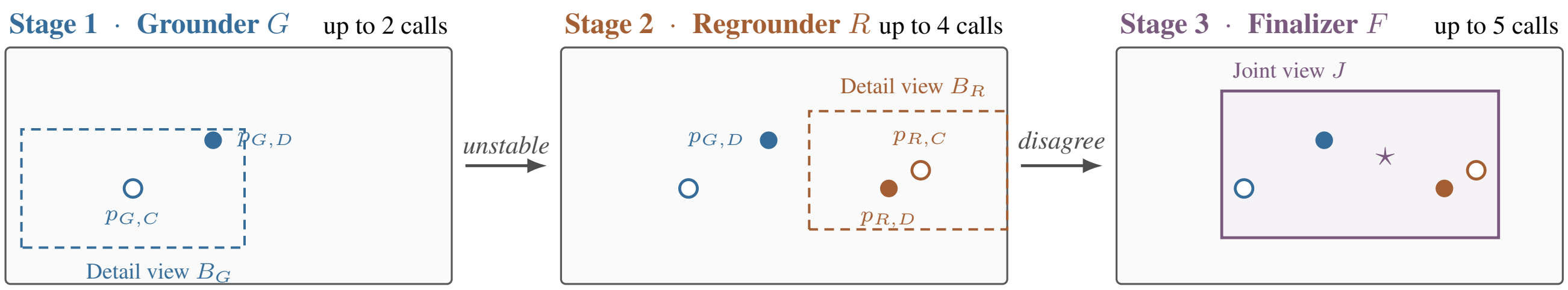
\begin{figure*}[t]
\centering
\begin{tikzpicture}[x=1cm,y=1cm,>=Latex,font=\small]
\colorlet{gcol}{gcG}\colorlet{rcol}{gcR}\colorlet{fcol}{gcF}\colorlet{gtxt}{gcGk}
\tikzset{screen/.style={draw=black!65,fill=black!2,rounded corners=1.5pt,line width=0.6pt},
  ctx/.style={circle,draw,line width=0.9pt,minimum size=5.5pt,inner sep=0pt,fill=white},
  det/.style={circle,fill,minimum size=5.5pt,inner sep=0pt},
  crop/.style={densely dashed,line width=0.75pt},
  stage/.style={font=\small\bfseries,anchor=south west,inner sep=1pt},
  note/.style={font=\footnotesize,align=left,anchor=north west,text width=4.85cm,inner sep=1pt},
  calls/.style={font=\footnotesize,anchor=south east,inner sep=1pt},
  flow/.style={-{Latex[length=2.2mm]},line width=0.9pt,black!70},
  lab/.style={font=\footnotesize\itshape,above}}
\begin{scope}[shift={(0,0)}]
 \node[stage,gtxt] at (0,2.65) {Stage 1 \;$\cdot$\; Grounder $G$};
 \draw[screen] (0,0) rectangle (4.9,2.6);
 \draw[crop,gcol] (0.175,0.40) rectangle (2.625,1.70);
 \node[font=\scriptsize,gtxt,anchor=north east] at (2.625,0.35) {Detail view $B_G$};
 \node[ctx,draw=gcol] at (1.40,1.05) {};
 \node[font=\scriptsize,gtxt,anchor=north] at (1.40,0.93) {$p_{G,C}$};
 \node[det,gcol] at (2.28,1.58) {};
 \node[font=\scriptsize,gtxt,anchor=west] at (2.42,1.58) {$p_{G,D}$};
 \node[note] at (0,-0.08) {Hollow $\circ$: Context; filled $\bullet$: Detail.\\
   \emph{Stop} if $s_G\le\tau_s$: return $p_{G,D}$.};
 \node[calls] at (4.9,2.65) {up to 2 calls};
\end{scope}
\draw[flow] (5.05,1.3) -- node[lab]{unstable} (5.95,1.3);
\begin{scope}[shift={(6.1,0)}]
 \node[stage,rcol] at (0,2.65) {Stage 2 \;$\cdot$\; Regrounder $R$};
 \draw[screen] (0,0) rectangle (4.9,2.6);
 \draw[crop,rcol] (2.725,0.60) rectangle (4.90,1.90);
 \node[font=\scriptsize,rcol,anchor=south east] at (4.80,1.94) {Detail view $B_R$};
 \node[ctx,draw=gcol] at (1.40,1.05) {};
 \node[det,gcol] at (2.28,1.58) {};
 \node[font=\scriptsize,gtxt,anchor=east] at (2.15,1.58) {$p_{G,D}$};
 \node[ctx,draw=rcol] at (3.95,1.25) {};
 \node[font=\scriptsize,rcol,anchor=south] at (3.95,1.38) {$p_{R,C}$};
 \node[det,rcol] at (3.60,1.05) {};
 \node[font=\scriptsize,rcol,anchor=north] at (3.60,0.92) {$p_{R,D}$};
 \node[note] at (0,-0.08) {$R$'s Context--Detail pair; $p_D^\star$ is the selected Detail point.
   \emph{Stop} on Detail agreement ($\tau_D$): return $p_D^\star$; otherwise on Context agreement ($\tau_C$): return $p_{m^\star,C}$.};
 \node[calls] at (4.9,2.65) {up to 4 calls};
\end{scope}
\draw[flow] (11.15,1.3) -- node[lab]{disagree} (12.05,1.3);
\begin{scope}[shift={(12.2,0)}]
 \node[stage,fcol] at (0,2.65) {Stage 3 \;$\cdot$\; Finalizer $F$};
 \draw[screen] (0,0) rectangle (4.9,2.6);
 \draw[fcol,line width=0.9pt,fill=fcol!7] (1.155,0.50847) rectangle (4.195,2.12153);
 \node[font=\scriptsize,fcol,anchor=south west] at (1.155,2.14) {Joint view $J$};
 \node[ctx,draw=gcol] at (1.40,1.05) {};
 \node[det,gcol] at (2.28,1.58) {};
 \node[ctx,draw=rcol] at (3.95,1.25) {};
 \node[det,rcol] at (3.60,1.05) {};
 \node[fcol,font=\large] at (2.95,1.40) {$\star$};
 \node[note] at (0,-0.08) {Only if both agreement tests fail, Context and Detail locations define $J$ (Eq.~\eqref{eq:joint}). $F$ returns a fresh prediction ($\star$).};
 \node[calls] at (4.9,2.65) {up to 5 calls};
\end{scope}
\end{tikzpicture}
\caption{\textbf{GroundCascade's two modes.} Selective uses Stages 1--2: stop when the Grounder is stable; otherwise acquire the Regrounder and test Detail, then tighter Context agreement. Full adds Stage 3, an unmarked joint view, only if both comparisons fail. Markers and outlines illustrate predicted locations, not model inputs; call labels are cumulative upper bounds because invalid Context skips Detail.}
\label{fig:method}
\end{figure*}

\gc controls second-model calls using the \emph{Grounder}'s Context--Detail agreement (Fig.~\ref{fig:method}). \mbox{It returns Detail} when the predictions agree closely and otherwise obtains a Context--Detail pair from the \emph{Regrounder}. Agreement between Detail predictions returns the more self-consistent one; when Detail disagrees, close Context agreement preserves the more self-consistent model's full-screen point. These two stages form the Selective variant; in its Full mode, \gc adds a \emph{Finalizer} on an unmarked joint crop only when both tests fail. Decisions use screen-space distances, without training, model confidences, or marked candidates.

\gc reaches \V{sspi8}{gc}\% on ScreenSpot-Pro (SSP) using Indeed-UI-32B/8B/32B, a 2.66-point gain over the stronger standalone two-pass grounder. GroundCascade tops the official SSP leaderboard~\cite{screenspot_leaderboard}. \mbox{With Qwen3.6-27B}/\allowbreak\mbox{3.8-27B}/\allowbreak\mbox{3.8-27B}, it reaches \V{mixosg}{gc}\% on the 510 visible-target requests of OSWorld-G-Refined (OSG-R), a 2.35-point gain. On UI-Vision Element Grounding (UIV), \gc reaches \V{mixuiv}{gc}\% category-balanced accuracy; omitting the Finalizer improves it slightly to \V{mixuiv}{sel}\% in Selective mode, 2.56 points above standalone Two-pass. The OSG-R and UIV results exceed the published comparisons with Hunmin-CUA and Qwen-UI-Agent, respectively (Fig.~\ref{fig:teaser}, Sec.~\ref{sec:main}). The acquisition gate outperforms stratified random acquisition at the same executed Regrounder-call budget on all three evaluations (Sec.~\ref{sec:selectivity}). Table~\ref{tab:main} presents the matched model-invocation costs.

The Regrounder need not be stronger in isolation: I8 trails I32 on SSP, and Q38 trails Q36 on OSG-R and UIV, yet selective acquisition improves the Grounder's outcomes. The Context return can repair zoom-induced losses; finalization adds 18 correct SSP answers and two OSG-R answers, with five fewer on UIV. Our error analysis separates missed escalation from the loss of correct predictions already acquired (Sec.~\ref{sec:analysis}).

Our contributions are:
\begin{itemize}
\item A training-free controller for selective second Context--Detail acquisition, corroborated Context preservation, and optional fresh unmarked joint-view prediction.
\item Matched-budget evidence that selective acquisition concentrates useful corrections over stratified random; analysis of Context preservation and conditional finalization.
\end{itemize}

\section{Related Work}
\label{sec:related}
\smallhead{Zoom and regional refinement.}
Iterative zooming and region search increase the effective resolution available for small targets~\cite{vtts,zoomclick,drsgui}; region ranking and visual-token processing refine how crops are used~\cite{rankground,reprefill}. BAMI and FastFocus combine coarse-to-fine views with candidate selection or verification~\cite{bami,fastfocus}, and GUI-Lens chooses crops for a general-purpose VLM~\cite{guilens}. Zooming can also discard needed context~\cite{zoomclick}. UI-Zoomer decides when to zoom from sampled consensus and token confidence~\cite{uizoomer}. MVP aggregates attention-guided views~\cite{mvp}; we evaluate its released pipeline with the same checkpoint. AutoFocus separates prediction, verification, and aggregation during active visual search~\cite{autofocus}; Chain-of-Ground, GMS, and MHVF likewise use feedback, regional specialization, or hypothesis verification~\cite{chainofground,gms,mhvf}.

\smallhead{Confidence, agreement, and verification.}
Zoom Consistency~\cite{zoomconsistency} uses consistency to route between two acquired two-pass pipelines and discusses early stopping in further zoom steps. GroundCascade instead uses the first pair to decide whether to acquire the second, with a separate, tighter test for preserving full-screen predictions. The matched unconditional-acquisition control is not a reproduction of Zoom Consistency's router. SafeGround calibrates sampled spatial dispersion for deferral~\cite{safeground}; PACE uses coordinate-token entropy~\cite{pace}, and ARGUS benchmarks grounding uncertainty~\cite{argus}. Unlike learned reflection~\cite{lookagain,copc}, GroundCascade's Finalizer makes an ordinary grounding prediction with the original instruction.

\smallhead{Model cascades.}
Cascades allocate computation selectively in detection~\cite{violajones}, neural networks~\cite{wisdom}, and language models~\cite{frugalgpt,motcascade}. Deferral theory distinguishes confidence in the current predictor from its performance relative to a downstream model~\cite{cascade_deferral}. Adaptive VLM Routing and LEASH study escalation in computer-use settings~\cite{avr,leash}. GroundCascade routes using only two-pass coordinates, without token probabilities or hidden states.

\section{GroundCascade}
\label{sec:method}
Let $I$ be a screenshot of width $W$ and height $H$, and let $q$ be an instruction. A grounding model $M$ maps $(I,q)$ to a point or an invalid output. Valid points use zero-based screenshot coordinates, $0\le x<W$ and $0\le y<H$. We use three views. The \emph{Context view} is the full screenshot, giving $p_{M,C}=M(I,q)$. The \emph{Detail view} is a crop $B_M$ of nominally half the screen width and height, centered on the Context point and clipped to the screenshot. Its prediction $p_{M,D}$ is mapped back to screenshot coordinates. An invalid Context output skips the Detail call and leaves $p_{M,D}$ invalid. The \emph{Joint view} is defined below. The three roles, Grounder $G$, Regrounder $R$, and Finalizer $F$, need not be distinct checkpoints. We evaluate I32/I8/I32 on SSP and Q36/Q38/Q38 on OSG-R and UIV (Sec.~\ref{sec:setup}); Sec.~\ref{sec:allocation} also tests 8B Grounders on SSP.

\smallhead{Screen-space displacement.}
All decisions compare mapped points with the axis-normalized $L_\infty$ distance
\begin{equation}
 d(a,b)=\max\left\{\frac{|a_x-b_x|}{W},\ \frac{|a_y-b_y|}{H}\right\}.
 \label{eq:distance}
\end{equation}
Threshold $\tau$ limits horizontal and vertical shifts to $\tau W$ and $\tau H$; invalid points have $d=\infty$.

Model self-displacement $s_M=d(p_{M,C},p_{M,D})$ uses screenshot coordinates, independent of crop position. It is related to zoom consistency~\cite{zoomconsistency}, which measures displacement from the crop center; ours uses axis-normalized $L_\infty$ distance to the original Context point, which can differ when the crop is clipped.

\smallhead{Stage 1: Grounder.}
$G$ produces its Context--Detail pair. If both predictions are valid and $s_G\le\tau_s$, the cascade returns $p_{G,D}$.

\smallhead{Stage 2: Regrounder.}
Otherwise, $R$ produces its own Context--Detail pair. We select the model with smaller self-displacement and its Detail candidate,
\begin{equation}
 p_D^\star=p_{m^\star,D},\qquad m^\star=\argmin_{M\in\{G,R\}} s_M,
 \label{eq:select}
\end{equation}
with ties broken in favor of $G$. If $p_D^\star$ is invalid, replace it with the other model's Detail prediction if valid. The Regrounder first checks Detail agreement: if $d(p_{G,D},p_{R,D})\le\tau_D$, it returns $p_D^\star$. Otherwise, if $d(p_{G,C},p_{R,C})\le\tau_C$, it returns $p_{m^\star,C}$. Because returning a Context prediction overrides the Detail-based choice, the Context-agreement test requires stronger cross-model agreement than the Detail-agreement test. Invalid points cannot satisfy an agreement test, and returned coordinates are never averaged.

\smallhead{Stage 3: Finalizer.}
The two-stage Selective variant stops here, returning $p_D^\star$ if neither agreement test passes. In Full mode, \gc instead makes one last prediction from a joint view in these unresolved cases, allowing a broader observation when Detail crops lose context. Let $\mathcal P$ contain the valid points among $p_{G,C},p_{G,D},p_{R,C},p_{R,D}$. If $\mathcal P$ is empty, $J$ is the full screenshot. Otherwise, let $x_-,x_+,y_-,y_+$ be the coordinate extrema and define
\begin{equation}
 \begin{aligned}
 \delta&=\max\left\{\frac{x_+-x_-}{W},\frac{y_+-y_-}{H}\right\},\\
 \rho&=\min\{1,\ \max\{0.5,\delta+0.1\}\}.
 \end{aligned}
 \label{eq:joint}
\end{equation}
The joint crop $J$ has side lengths $\rho W,\rho H$, rounded to pixels, and is centered at the bounding-box midpoint $((x_-+x_+)/2,(y_-+y_+)/2)$. Unlike the clipped Detail crop, $J$ is shifted inside the screenshot to preserve its rounded dimensions and retain all valid in-screen points up to pixel rounding. The two agreement comparisons determine whether $F$ is needed; all available valid points determine its view.

$F$ receives the unmarked view $J$ and the original instruction, without candidate coordinates. Its prediction $p_{F,J}$ is mapped back to the screenshot. The cascade returns the first valid point in $(p_{F,J},p_D^\star,p_{G,C},p_{R,C})$, or an invalid output if none exists.

\smallhead{Budget and parameters.}
Let $r,f\in\{0,1\}$ indicate whether the Regrounder and Finalizer are invoked. The Grounder uses at most two calls, the Regrounder at most two when invoked, and the Finalizer one. Thus, the number of executed calls per request, $L$, satisfies
\begin{equation}
 L\le 2+2r+f\le 5.
 \label{eq:budget}
\end{equation}
We report executed calls, both in total and per checkpoint. The thresholds used throughout are $\tau_s=0.01$ (self-stability), $\tau_D=0.04$ (Detail agreement), and $\tau_C=0.005$ (Context agreement), all in the normalized metric of Eq.~\eqref{eq:distance}. Supplementary Sec.~\ref{sec:s_impl} gives the exact transforms.

\section{Experiments}
\label{sec:experiments}
\begin{table*}[t]
\centering\small
\setlength{\tabcolsep}{3.5pt}
\begin{tabular*}{\textwidth}{@{\extracolsep{\fill}}llrrrrrrr@{}}
\toprule
 & & \multicolumn{2}{c}{Standalone accuracy (\%)} & \multicolumn{2}{c}{GroundCascade accuracy (\%)} & \multicolumn{3}{c}{Inference cost} \\
\cmidrule(lr){3-4}\cmidrule(lr){5-6}\cmidrule(lr){7-9}
$G/R/F$ & Benchmark & Context & Two-pass & Selective & Full & Selective & Full & Saved (\%) \\
\midrule
I32/I8/I32 & SSP & \V{sspi8}{ctx} & \V{sspi8}{zoom} & \V{sspi8}{sel} & \textbf{\V{sspi8}{gc}} & \V{sspi8}{selCalls} & \V{sspi8}{calls} & \V{sspi8}{fullSaving} \\
\midrule
\multirow{2}{*}{Q36/Q38/Q38}
 & OSG-R & \V{mixosg}{bestctx} & \V{mixosg}{detail} & \V{mixosg}{sel} & \textbf{\V{mixosg}{gc}} & \V{mixosg}{callsnof} & \V{mixosg}{calls} & \V{mixosg}{saved} \\
 & UIV & \V{mixuiv}{bestctx} & \V{mixuiv}{detail} & \textbf{\V{mixuiv}{sel}} & \V{mixuiv}{gc} & \V{mixuiv}{callsnof} & \V{mixuiv}{calls} & \V{mixuiv}{saved} \\
\bottomrule
\end{tabular*}
\caption{\textbf{Accuracy and inference cost on three benchmarks.} Model allocations ($G/R/F$) are specified by row. Context and Two-pass show the stronger standalone model per benchmark. Selective uses Stages~1--2; Full additionally invokes the Finalizer on unresolved requests. Inference cost is calls/query for each mode, followed by Full's savings relative to unconditional acquisition of both two-pass models. UIV accuracy is category-balanced; bold marks the highest accuracy in each row.}
\label{tab:main}
\end{table*}

\subsection{Setup}
\label{sec:setup}
\smallhead{Benchmarks.}
We evaluate ScreenSpot-Pro (SSP; \V{sspi8}{n} professional-interface requests)~\cite{screenspot}, OSWorld-G-Refined (OSG-R; \V{mixosg}{n} visible-target requests, excluding the 54 absent-target cases in the original 564-item benchmark)~\cite{osworld}, and UI-Vision Element Grounding (UIV; \V{mixuiv}{n} instructions over 1,181 screenshot files containing 1,156 distinct images)~\cite{uivision}. A point must fall inside its target region; invalid outputs are errors. UIV scores the unweighted mean of Basic, Functional, and Spatial accuracies; SSP and OSG-R use per-request accuracy.

\smallhead{Models and controls.}
We evaluate Indeed-UI-32B (I32), Indeed-UI-8B (I8)~\cite{indeedui}, KV-Ground-8B (K8)~\cite{kvground}, Qwen3.6-27B (Q36)~\cite{qwen36card}, and Qwen3.8-27B (Q38)~\cite{qwen38card}. Configurations denote $G/R/F$; Selective does not invoke $F$. Table~\ref{tab:main} reports I32/I8/I32 on SSP and Q36/Q38/Q38 on OSG-R and UIV. We test additional Indeed model orders on SSP, compare with the released MVP~\cite{mvp} pipeline, and report a UI-Venus-2~\cite{venus2} interface check in the supplement. Matched controls reuse recorded predictions and views.

\smallhead{Evaluation protocol.}
Paired 95\% percentile accuracy confidence intervals (CIs) use 10,000 bootstrap resamples: by request for SSP and by distinct image for OSG-R and UIV. The CIs are conditional on the reported configurations and do not account for configuration-selection uncertainty. Calls count executed model invocations, not tokens, runtime, or FLOPs (supplementary Sec.~\ref{sec:s_metrics}).

\begin{figure*}[!t]
\centering
\input{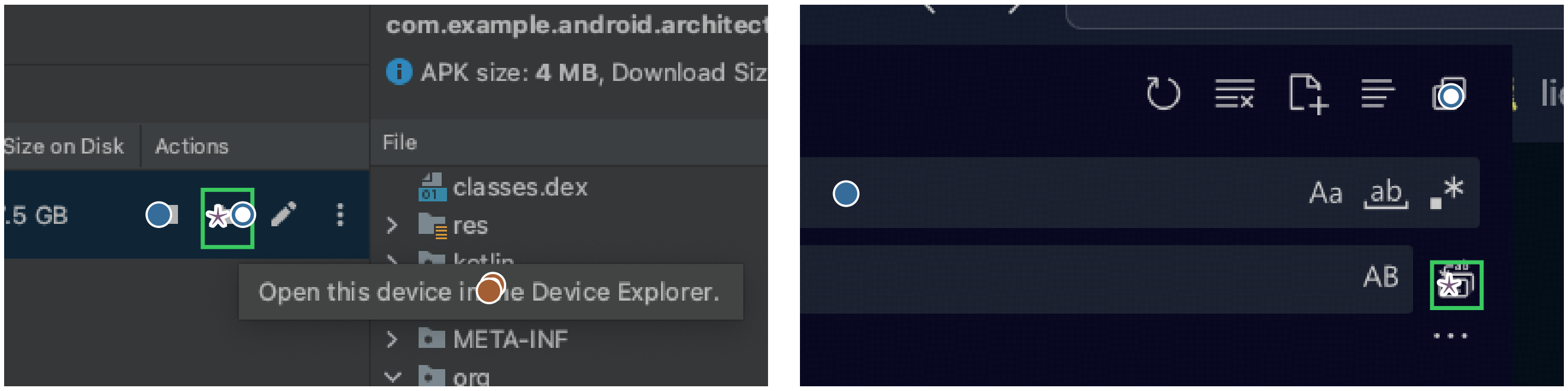}

\begin{minipage}[t]{.49\textwidth}
\centering
{\footnotesize ``open the device in the device explorer''\par}
\vspace{5pt}
\GCAndroidPanel
\par\vspace{2pt}
{\small (a) Zoom error repaired\par}
\end{minipage}\hfill
\begin{minipage}[t]{.49\textwidth}
\centering
{\footnotesize ``Replace \textquotedblleft cat\textquotedblright{} with
\textquotedblleft dog\textquotedblright{} in all files.''\par}
\vspace{5pt}
\GCVSCodePanel
\par\vspace{2pt}
{\small (b) New location found\par}
\end{minipage}

\par\vspace{3pt}
\centerline{%
\begin{tikzpicture}[x=1pt,y=-1pt,inner sep=0pt,outer sep=0pt]
\path[use as bounding box] (0,0) rectangle (\textwidth,12);
\begin{scope}[shift={(3,-167.5)}]
\draw[draw=gcg,fill=white,line width=1pt] (102.1,173.5) circle (3.5pt);
\node[anchor=west,font=\footnotesize] at (109.1,173.5) {G Context};
\draw[draw=gcg,fill=gcg,line width=1pt] (159.2,173.5) circle (3.5pt);
\node[anchor=west,font=\footnotesize] at (166.2,173.5) {G Detail};
\draw[draw=gcr,fill=white,line width=1pt] (209.2,173.5) circle (3.5pt);
\node[anchor=west,font=\footnotesize] at (216.2,173.5) {R Context};
\draw[draw=gcr,fill=gcr,line width=1pt] (265.9,173.5) circle (3.5pt);
\node[anchor=west,font=\footnotesize] at (272.9,173.5) {R Detail};
\node[font=\large,text=gcf] at (315.0,173.5) {$\star$};
\node[anchor=west,font=\footnotesize] at (321.0,173.5) {Finalizer};
\draw[draw=gctarget,line width=1.3pt] (360.4,170) rectangle (371.4,177);
\node[anchor=west,font=\footnotesize] at (375.4,173.5) {Target};
\end{scope}
\end{tikzpicture}%
}
\caption{\textbf{ScreenSpot-Pro recoveries.} (a) The Finalizer repairs a zoom-induced error. (b) It finds a target missed by all four upstream predictions (the $R$ points lie outside this excerpt). Prediction circles, Finalizer stars, and target boxes are evaluation overlays, not model inputs.}
\label{fig:qualitative}
\end{figure*}

\subsection{Main results}
\label{sec:main}
Table~\ref{tab:main} compares the three reported allocations with the stronger standalone Context and Two-pass models on each benchmark.
On SSP, \gc (I32/I8/I32) reaches \V{sspi8}{gc}\% (\V{sspi8}{gcN}/\V{sspi8}{n}), compared with \V{sspi8}{zoom}\% for standalone Two-pass: \V{sspi8}{dzoomN} additional correct answers, or $+2.66$ points (paired 95\% CI \V{sspi8}{cizoom} points). Selective reaches \V{sspi8}{sel}\%; \gc uses \V{sspi8}{calls} calls/query, \V{sspi8}{fullSaving}\% fewer than unconditional acquisition.

On OSG-R, \gc (Q36/Q38/Q38) reaches \V{mixosg}{gc}\% (\V{mixosg}{correct}/\V{mixosg}{n}), compared with \V{mixosg}{detail}\% for standalone Q36 Two-pass: 12 additional correct answers, or $+2.35$ points (image-clustered 95\% CI [+\V{mixosg}{detailcilo},\,+\V{mixosg}{detailcihi}] points). Selective reaches \V{mixosg}{sel}\%; \gc uses \V{mixosg}{calls} calls/query, saving \V{mixosg}{saved}\%.

On UIV, \gc (Q36/Q38/Q38) reaches \V{mixuiv}{gc}\% category-balanced accuracy (\V{mixuiv}{correct}/\V{mixuiv}{n} correct, \V{mixuiv}{pooled}\% pooled), a $+2.48$-point gain over standalone Q36 Two-pass (\V{mixuiv}{detail}\%; image-clustered 95\% CI [+\V{mixuiv}{detailcilo},\,+\V{mixuiv}{detailcihi}] points). Omitting the Finalizer yields \V{mixuiv}{sel}\% in Selective mode, a $+2.56$-point gain over the same baseline (95\% CI [+\V{mixuiv}{selcilo},\,+\V{mixuiv}{selcihi}] points). Q38 alone scores \V{mixuiv}{rdetail}\% in Two-pass, but complements Q36.

\smallhead{Official leaderboard and published comparisons.}
Figure~\ref{fig:teaser} compares \gc with selected publicly reported GUI-focused systems. On SSP, the strongest non-GroundCascade leaderboard entry is Indeed-UI-32B-zoomin at 82.73\%~\cite{indeedui,screenspot_leaderboard}. On OSG-R, \gc reaches \V{mixosg}{gc}\% versus Hunmin-CUA's 86.8\% on the same 510 visible-target items~\cite{hunmin_cua}. Four other published scores are shown only as conditional \emph{upper bounds} under the 564-item micro-accuracy interpretation, not as measured 510-item results (supplementary Sec.~\ref{sec:s_metrics})~\cite{venus2,qwenuiagent,venus15}. On UIV, \gc reaches \V{mixuiv}{gc}\% versus the published 70.0\% for Qwen-UI-Agent-27B~\cite{qwenuiagent}, whose averaging convention is unspecified; Selective reaches \V{mixuiv}{sel}\% (Tab.~\ref{tab:main}). GroundNext's training applications overlap UI-Vision's~\cite{groundnext}. External checkpoints and inference protocols differ. Proprietary general-purpose systems have reported higher SSP scores (GPT-6 Astra: 92.7\%; GUI-Lens with GPT-5.5: 87.9\%) and are outside the specialist plot~\cite{gpt6astra,guilens}. Table~\ref{tab:main} presents the controlled within-allocation comparisons.

Across SSP, OSG-R, and UIV, unconditional two-model Detail selection scores \V{sspi8}{ao}\%, \V{mixosg}{ao}\%, and \V{mixuiv}{ao}\%; \gc scores \V{sspi8}{gc}\%, \V{mixosg}{gc}\%, and \V{mixuiv}{gc}\% with 2.42--2.73 calls/query (31--39\% fewer). On UIV, the accuracy difference is small (paired CI [\V{mixuiv}{cilo},\,+\V{mixuiv}{cihi}] points). Figure~\ref{fig:budget} isolates the value of selective acquisition.

On SSP, the released MVP pipeline~\cite{mvp} matches I32 Two-pass at \V{mvp}{acc}\% using \V{mvp}{calls} I32 calls/query. \gc gains \V{mvp}{diffp} points (paired 95\% CI \V{mvp}{ci} points) with \V{sspi8}{calls} calls/query, including its I8 Regrounder. MVP uses only I32, whereas \gc also uses I8; image and generation budgets differ, so the comparison concerns executed calls rather than equal compute.

\subsection{Model identity and order}
\label{sec:allocation}

\begin{table}[!b]
\centering\small
\setlength{\tabcolsep}{4.2pt}
\begin{tabular*}{\columnwidth}{@{\extracolsep{\fill}}lrrrrrr@{}}
\toprule
 & \multicolumn{3}{c}{I32 first: I32/8B/I32} & \multicolumn{3}{c}{8B first: 8B/I32/I32} \\
\cmidrule(lr){2-4}\cmidrule(lr){5-7}
8B model & Acc. & Calls & I32 & Acc. & Calls & I32 \\
\midrule
I8 & \textbf{\V{sspi8}{gc}} & \V{sspi8}{calls} & \V{sspi8}{g_calls} & \V{ord}{i8acc} & \V{ord}{i8calls} & \textbf{\V{ord}{i8large}} \\
K8 & \V{sspk8}{gc} & \V{sspk8}{calls} & \V{sspk8}{g_calls} & \V{ord}{k8acc} & \V{ord}{k8calls} & \V{ord}{k8large} \\
\bottomrule
\end{tabular*}
\caption{\textbf{Model identity and order on SSP} (accuracy, \%). Calls and I32 are mean total and 32B model invocations per request, respectively. Bold marks the highest accuracy and the fewest I32 calls.}
\label{tab:roles}
\end{table}

On SSP, all four allocations outperform their Grounder's standalone Two-pass baseline. I32/K8/I32 reaches \V{sspk8}{gc}\%; using I8 or K8 first adds \V{ord}{i8downzoom} and \V{ord}{k8downzoom} correct answers over their standalone baselines.

Putting an 8B model first switches the Grounder and Regrounder, while I32 remains the Finalizer (Tab.~\ref{tab:roles}). On SSP, I8/I32/I32 reaches \V{ord}{i8acc}\% using \V{ord}{i8large} I32 calls per request, versus \V{sspi8}{gc}\% and \V{sspi8}{g_calls} I32 calls for I32/I8/I32. This saves \V{ord}{i8saveflag}\% of I32 calls at a cost of \V{ord}{i8dflag} correct answers (paired CI for I8-first minus I32-first: \V{ord}{i8ciflag} points). K8-first follows the same trade-off (Tab.~\ref{tab:roles}). Both orders use about 2.5 calls per request; I32-first favors accuracy.

\subsection{Why selectivity helps}
\label{sec:selectivity}
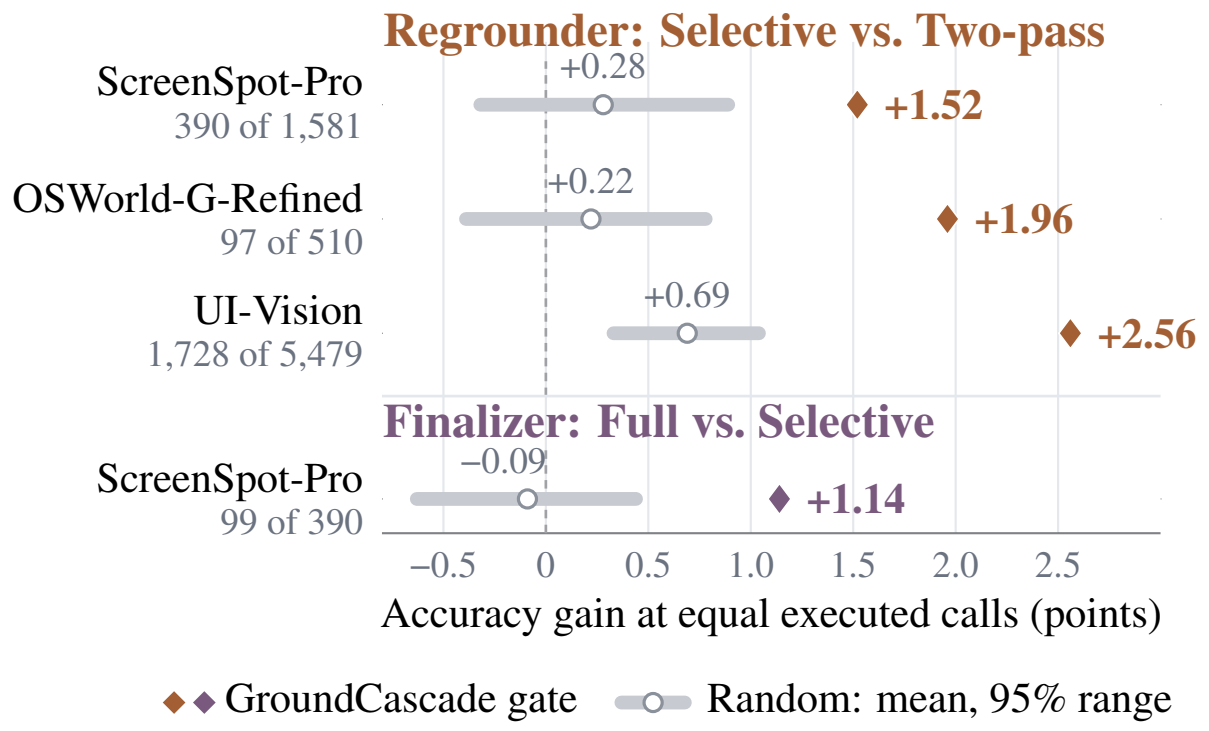
\begin{figure}[t]
\centering
\begingroup
\colorlet{gB}{gcG}\colorlet{gBk}{gcGk}\colorlet{gR}{gcR}\colorlet{gF}{gcF}
\definecolor{ink}{HTML}{2B2F36}\definecolor{muted}{HTML}{6B7280}
\definecolor{gridc}{HTML}{E4E7EB}\definecolor{rnd}{HTML}{9AA1AB}
\newcommand{\fsS}{\scriptsize}
\newcommand{\fsT}{\footnotesize}
\newcommand{\fsH}{\small}
\newcommand{\fsLeg}{\footnotesize}

\newcommand{\row}[8]{%
  \draw[rnd!55, line width=2.6pt, line cap=round] (axis cs:#2,#1) -- (axis cs:#4,#1);
  \addplot[only marks, mark=*, mark size=1.7pt, draw=rnd!90!black, fill=white, line width=0.6pt]
    coordinates {(#3,#1)};
  \addplot[only marks, mark=diamond*, mark size=2.9pt, color=#6, draw=white, line width=0.3pt]
    coordinates {(#5,#1)};
  \node[font=\fsT\bfseries, text=#6, anchor=west] at (axis cs:#5,#1) {\hspace{1.6pt}#7};
  \node[font=\fsS, text=muted, anchor=south] at (axis cs:#3,#1) {\raisebox{1.4pt}{#8}};
}
\begin{tikzpicture}
\begin{axis}[
  width=5.15cm, height=3.25cm, scale only axis, clip=false,
  xmin=-0.8, xmax=3.0, ymin=0.25, ymax=4.55,
  xtick={-0.5,0,0.5,1,1.5,2,2.5},
  xticklabels={\textminus0.5,0,0.5,1.0,1.5,2.0,2.5},
  ytick={4,3,2,0.55},
  yticklabels={{ScreenSpot-Pro\\[-1.5pt]{\fsS\color{muted}390 of 1,581}},
               {OSWorld-G-Refined\\[-1.5pt]{\fsS\color{muted}97 of 510}},
               {UI-Vision\\[-1.5pt]{\fsS\color{muted}1,728 of 5,479}},
               {ScreenSpot-Pro\\[-1.5pt]{\fsS\color{muted}99 of 390}}},
  yticklabel style={font=\fsT, align=right},
  xticklabel style={font=\fsS, text=muted},
  axis line style={draw=none}, axis x line*=bottom,
  x axis line style={draw=ink!60, line width=0.4pt},
  xmajorgrids, grid style={draw=gridc, line width=0.4pt},
  major tick length=0pt,
  xlabel={Accuracy gain at equal executed calls (points)},
  xlabel style={font=\fsT, yshift=3pt},
]
  \draw[ink!45, line width=0.5pt, dash pattern=on 1.6pt off 1.2pt] (axis cs:0,0.25) -- (axis cs:0,1.02);
  \draw[ink!45, line width=0.5pt, dash pattern=on 1.6pt off 1.2pt] (axis cs:0,1.45) -- (axis cs:0,4.38);
  \node[font=\fsH\bfseries, text=gR, anchor=south west, inner sep=0pt]
    at (axis cs:-0.8,4.42) {Regrounder: Selective vs.\ Two-pass};
  \node[font=\fsH\bfseries, text=gF, anchor=south west, inner sep=0pt]
    at (axis cs:-0.8,1.08) {Finalizer: Full vs.\ Selective};
  \draw[gridc, line width=0.5pt] (axis cs:-0.8,1.45) -- (axis cs:3.0,1.45);
  \row{4}{-0.32}{0.28}{0.89}{1.52}{gR}{+1.52}{+0.28}
  \row{3}{-0.39}{0.22}{0.78}{1.96}{gR}{+1.96}{+0.22}
  \row{2}{0.33}{0.69}{1.04}{2.56}{gR}{+2.56}{+0.69}
  \row{0.55}{-0.63}{-0.09}{0.44}{1.14}{gF}{+1.14}{\textminus0.09\hspace{9pt}}
\end{axis}
\begin{scope}[shift={(-1.55,-1.12)}, font=\fsLeg]
  \fill[gR] (0.10,0) -- ++(0.075,0.075) -- ++(0.075,-0.075) -- ++(-0.075,-0.075) -- cycle;
  \fill[gF] (0.30,0) -- ++(0.075,0.075) -- ++(0.075,-0.075) -- ++(-0.075,-0.075) -- cycle;
  \node[anchor=west, inner sep=1pt] (a) at (0.47,0) {GroundCascade gate};
  \draw[rnd!55, line width=2.6pt, line cap=round] ([xshift=7pt]a.east) -- ++(0.42,0);
  \draw[rnd!90!black, fill=white, line width=0.6pt] ([xshift={7pt+0.21cm}]a.east) circle (1.7pt);
  \node[anchor=west, inner sep=1pt] at ([xshift={10pt+0.42cm}]a.east)
    {Random: mean, 95\% range};
\end{scope}
\end{tikzpicture}
\endgroup
\caption{\textbf{Choosing which requests get a second opinion matters more than the extra calls.} Diamonds: accuracy gain from the GroundCascade gate. Gray: the same number of executed calls given to randomly chosen requests (mean and central 95\% range). Top: Regrounder acquisition (Selective \vs standalone Two-pass); random allocations are stratified by recorded call cost (and UIV category), and labels give the requests that received $R$. Bottom: Finalizer calls on SSP (Full \vs Selective), drawn uniformly among the \V{erssp}{gGateDeferred} eligible requests. Both controls use 50,000 draws. SSP uses I32/I8/I32; OSG-R and UIV use Q36/Q38/Q38; UIV is category-balanced. Supplementary Tab.~\ref{tab:s_helper_budget} lists the values.}
\label{fig:budget}
\end{figure}

Figure~\ref{fig:budget} isolates the advantage of selecting which requests receive the second model.

\smallhead{Acquiring the second model.}
At the same executed Regrounder-call budgets, stratified random acquisition provides smaller mean gains (Fig.~\ref{fig:budget}). On SSP, acquiring $R$ on every early exit instead loses six correct answers. The gate therefore concentrates complementary predictions on requests with greater intervention value.

\begin{figure}[t]
\centering
\begingroup
\colorlet{gB}{gcG}\colorlet{gBk}{gcGk}\colorlet{gR}{gcR}\colorlet{gF}{gcF}
\colorlet{tB}{gcG!50!white}
\definecolor{ink}{HTML}{2B2F36}\definecolor{muted}{HTML}{6B7280}\definecolor{gridc}{HTML}{E4E7EB}
\newcommand{\fsS}{\scriptsize}
\newcommand{\fsT}{\footnotesize}
\newcommand{\fsV}{\footnotesize}
\newcommand{\fsLeg}{\footnotesize}

\def\u{1.56}\def\xo{2.29}\def\bh{0.36}
\newcommand{\X}[1]{\fpeval{\xo + \u*(#1)}}
\begin{tikzpicture}[x=1cm, y=1cm]
  \foreach \t in {0,0.5,1,1.5,2,2.5,3} {
    \draw[gridc, line width=0.4pt] (\X{\t},-0.05) -- (\X{\t},2.62);
    \node[font=\fsS, text=muted, anchor=north] at (\X{\t},-0.07) {\t};
  }
  \draw[ink!60, line width=0.45pt] (\X{0},-0.05) -- (\X{0},2.62);
  \node[font=\fsT, anchor=north] at (\X{1.5},-0.36) {Gain (percentage points)};

  \newcommand{\row}[9]{%
    \node[font=\fsT, anchor=east, align=right] at (\xo-0.18,#1)
      {#2\\[-1.5pt]{\fsS\color{muted}#3}};
    \fill[gR] (\X{0},#1-\bh/2) rectangle (\X{#4},#1+\bh/2);
    \node[font=\fsS, text=white] at (\X{#4/2},#1) {#7};
    \fill[tB] (\X{#4},#1-\bh/2) rectangle (\X{#4+#5},#1+\bh/2);
    #9%
  }
  \row{2.18}{ScreenSpot-Pro}{I32/I8/I32}{1.4548}{0.0633}{1.1385}{+23}{+1}{%
    \fill[gF] (\X{1.5181},2.18-\bh/2) rectangle (\X{2.6566},2.18+\bh/2);
    \node[font=\fsS, text=white] at (\X{2.0874},2.18) {+18 (27 / 9)};
    \draw[ink, line width=0.9pt] (\X{2.6566},2.18-\bh/2-0.03) -- (\X{2.6566},2.18+\bh/2+0.03);
    \node[font=\fsV\bfseries, text=gBk, anchor=west] at (\X{2.6566}+0.05,2.18) {+2.66};
    \node[font=\fsS, text=gBk, anchor=south] at (\X{1.4865},2.18+\bh/2) {+1};}
  \row{1.30}{OSWorld-G-Refined}{Q36/Q38/Q38}{1.9608}{0}{0.3922}{+10}{0}{%
    \fill[gF] (\X{1.9608},1.30-\bh/2) rectangle (\X{2.3529},1.30+\bh/2);
    \node[font=\fsS, text=gF, anchor=south] at (\X{2.1569},1.30+\bh/2) {+2 (3 / 1)};
    \node[font=\fsV\bfseries, text=gBk, anchor=west] at (\X{2.3529}+0.05,1.30) {+2.35};
    \draw[ink, line width=0.9pt] (\X{2.3529},1.30-\bh/2-0.03) -- (\X{2.3529},1.30+\bh/2+0.03);}
  \row{0.42}{UI-Vision}{Q36/Q38/Q38}{2.3544}{0.1825}{-0.0913}{+129}{+10}{%
    \path[fill=white, postaction={pattern=north east lines, pattern color=gF}]
      (\X{2.4456},0.42-\bh/2) rectangle (\X{2.5369},0.42+\bh/2);
    \draw[gF, line width=0.4pt] (\X{2.4456},0.42-\bh/2) rectangle (\X{2.5369},0.42+\bh/2);
    \draw[ink, line width=0.9pt] (\X{2.4456},0.42-\bh/2-0.03) -- (\X{2.4456},0.42+\bh/2+0.03);
    \node[font=\fsS, text=gBk, anchor=south] at (\X{2.4456},0.42+\bh/2) {+10};
    \node[font=\fsS, text=gF, anchor=north] at (\X{2.4913},0.42-\bh/2) {\textminus5 (70 / 75)};
    \node[font=\fsV\bfseries, text=gBk, anchor=west] at (\X{2.5369}+0.05,0.42) {+2.45};}

  \begin{scope}[shift={(0.02,-0.98)}, font=\fsLeg]
    \fill[gR] (0,-0.065) rectangle ++(0.2,0.13);
    \node[anchor=west, inner sep=0pt, xshift=3pt] (a) at (0.2,0) {Detail selection};
    \fill[tB] ([xshift=7pt]a.east) ++(0,-0.065) rectangle ++(0.2,0.13);
    \node[anchor=west, inner sep=0pt] (b) at ([xshift={10pt+0.2cm}]a.east) {Context kept};
    \fill[gF] ([xshift=7pt]b.east) ++(0,-0.065) rectangle ++(0.2,0.13);
    \node[anchor=west, inner sep=0pt] at ([xshift={10pt+0.2cm}]b.east) {Finalizer net};
  \end{scope}
\end{tikzpicture}
\endgroup
\caption{\textbf{Where the gain comes from.} Stage contributions to the accuracy gain over the stronger standalone Two-pass model; annotations show net correct-answer counts. Allocations follow $G/R/F$ as in Tab.~\ref{tab:main}. Choosing between the two models' Detail points gives most of the gain on every benchmark; preserving corroborated Context points adds little; the Finalizer helps on SSP (\V{sspi8}{fRep} repairs, \V{sspi8}{fReg} regressions) and slightly hurts on UIV (\V{mixuiv}{fRep} repairs, \V{mixuiv}{fReg} regressions). Black ticks mark Full. UIV is pooled over its \V{mixuiv}{n} requests here (category-balanced gain: $+$2.48).}
\label{fig:gain}
\end{figure}
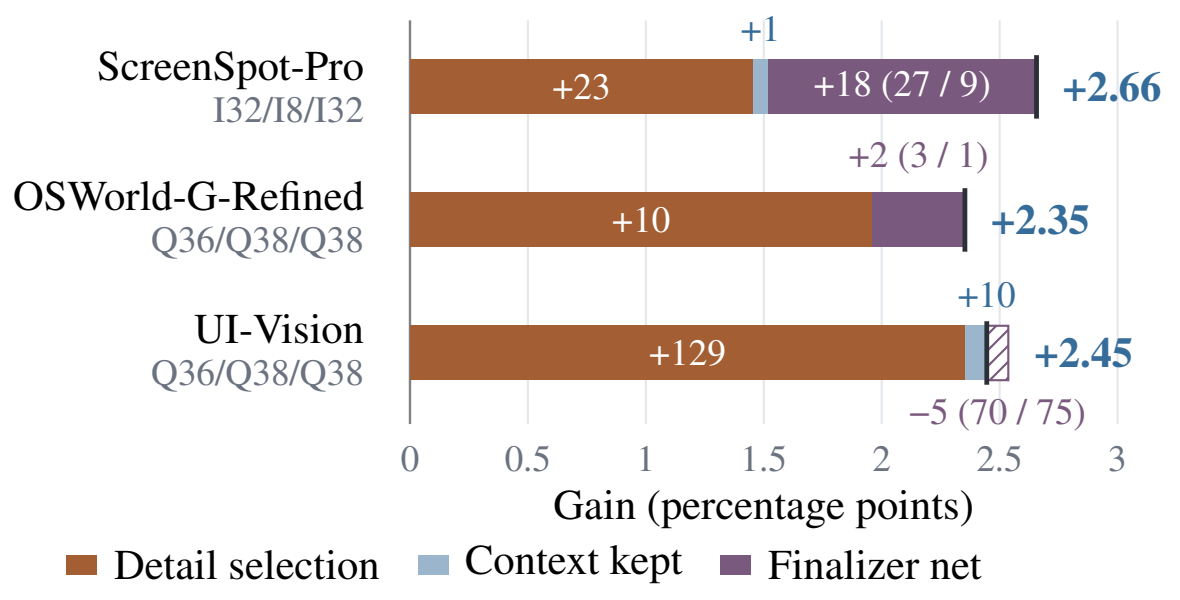

\smallhead{Resolving disagreement.}
On SSP, the Context-agreement branch is taken on \V{sspi8}{ccN} requests, yielding a net gain of \V{sspi8}{ccNet} correct answer over the preferred Detail candidate. Selective adds \V{sspi8}{stage2Gain} correct answers over standalone Two-pass; Full's \V{sspi8}{fnF} Finalizer calls add another \V{sspi8}{fNet}. For Qwen OSG-R, selective acquisition adds ten correct answers, followed by two more from the \V{mixosg}{fReq} Finalizer calls. On Qwen UIV, Selective adds 139 net correct requests over standalone Two-pass; the \V{mixuiv}{fReq} Finalizer calls then reduce the correct count by five. Figure~\ref{fig:gain} decomposes these gains by stage, and supplementary Tab.~\ref{tab:s_finalizer} gives repairs and regressions for each benchmark.

On SSP, Finalizer counterfactual predictions are available for all \V{erssp}{gGateDeferred} requests reaching $R$. Applying $F$ to all \V{ivssp}{FSkippedN} agreement exits would net $\V{ivssp}{FSkippedNet}$ correct answers at \V{ivssp}{FSkippedN} extra calls. Uniform \V{sspi8}{fnF}-call $F$ allocation yields a central 95\% net-change range $[\V{ivssp}{FUniformLow},+\V{ivssp}{FUniformHigh}]$, versus the gate's $+\V{ivssp}{FSelectedNet}$ (Fig.~\ref{fig:budget}, bottom).

\section{Error Detection and Recovery}
\label{sec:analysis}
We analyze routing, recovery, and remaining errors for the same three allocations reported in Table~\ref{tab:main}.

\subsection{Agreement as a routing signal}
\label{sec:signals}
Smaller Context-to-Detail displacement predicts greater reliability: using negative displacement to rank Grounder Detail correctness gives AUROC \V{sig}{allscreenlinf} on SSP. At the deployed threshold, the first gate escalates \V{erssp}{gGateTP} of \V{erssp}{gGateErrors} standalone Two-pass errors on SSP and \V{mixosg}{gGateTP} of \V{mixosg}{gGateErr} on Qwen OSG-R. Supplementary Tab.~\ref{tab:s_error_recovery} reports precision and recall across all three benchmarks. Escalation concentrates errors but also subjects some correct predictions to further inference.

\subsection{Recovery and preservation}
\label{sec:zoomhurts}
Zooming can turn correct predictions wrong~\cite{zoomclick,uizoomer}. On SSP, it spoils \V{zr}{sspi8breaks} correct Context predictions; the first gate flags \V{zr}{sspi8esc}, and the cascade restores \V{zr}{sspi8rec}. Selecting only between Detail predictions cannot directly recover a correct full-screen location lost through zooming. Stage~2 can instead preserve the more self-consistent model's Context prediction when both full-screen predictions agree within $\tau_C$ along each axis.

On SSP, the Context-agreement branch returns 13 correct predictions on \V{sspi8}{ccN} requests, versus 12 from the preferred Detail candidate. Across these benchmarks, Context preservation contributes less than the Detail-based routing decisions. Detail itself repairs \V{erssp}{zoomRepairs} Context errors, of which \V{erssp}{zoomRepairsKept} remain correct after \gc. All \V{erssp}{cascadeRegressions} cascade regressions relative to standalone Two-pass lose a successful zoom correction. The controller must balance recovery of Context with preservation of useful Detail.

\subsection{Where errors remain}
\label{sec:stability}
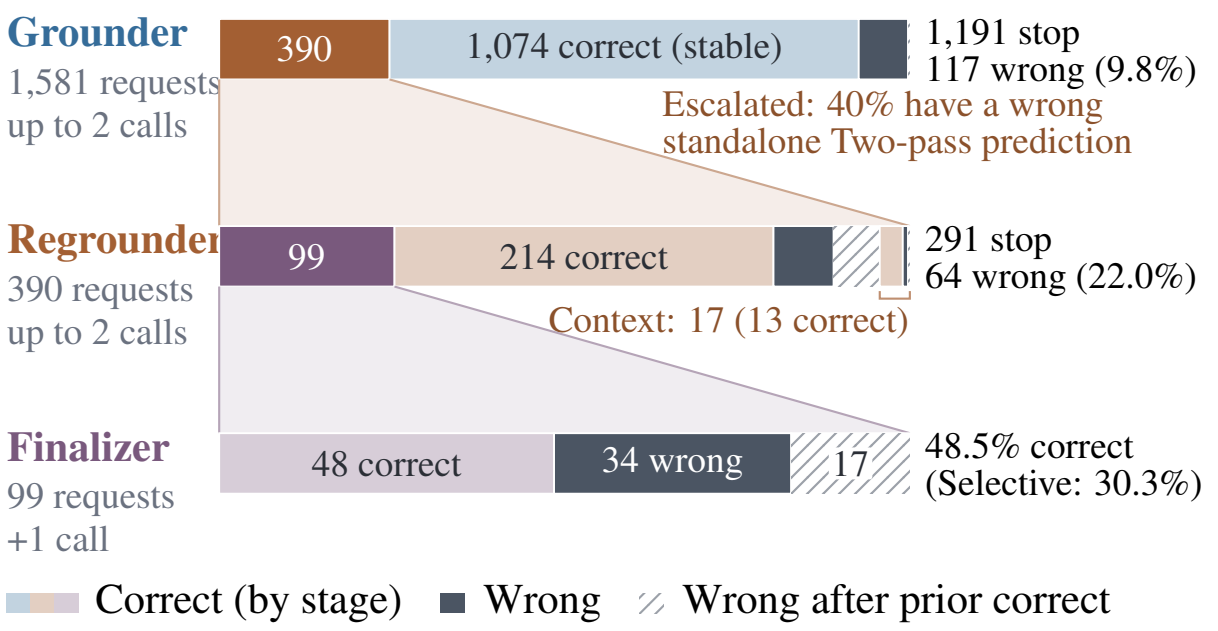
\begin{figure}[t]
\centering
\begingroup
\colorlet{gB}{gcG}\colorlet{gBk}{gcGk}\colorlet{gR}{gcR}\colorlet{gF}{gcF}
\colorlet{tB}{gcGt}\colorlet{tR}{gcRt}\colorlet{tF}{gcFt}
\definecolor{wrong}{HTML}{4B5563}\definecolor{ink}{HTML}{2B2F36}\definecolor{muted}{HTML}{6B7280}
\newcommand{\fsS}{\scriptsize}
\newcommand{\fsL}{\footnotesize}
\newcommand{\fsLeg}{\footnotesize}
\tikzset{
  seg/.style={draw=white, line width=0.35pt},
  recov/.style={seg, fill=white, postaction={pattern=north east lines, pattern color=wrong!55}},
  lab/.style={font=\fsS, inner sep=0.6pt},
}
\begin{tikzpicture}[x=0.94cm,y=1cm]
\node[anchor=north west, inner sep=0pt, align=left, font=\fsL\bfseries, text=gBk] at (0,3.7200) {Grounder\\[-0.5pt]{\fsS\mdseries\color{muted}1,581 requests}\\[-1.5pt]{\fsS\mdseries\color{muted}up to 2 calls}};
\path[seg, fill=gR] (1.5000,3.3000) rectangle (2.6989,3.7000);
\node[lab, text=white] at (2.0994,3.5000) {390};
\path[seg, fill=tB] (2.6989,3.3000) rectangle (6.0003,3.7000);
\node[lab, text=ink] at (4.3496,3.5000) {1,074 correct (stable)};
\path[seg, fill=wrong] (6.0003,3.3000) rectangle (6.3477,3.7000);
\path[recov] (6.3477,3.3000) rectangle (6.3600,3.7000);
\node[anchor=north west, inner sep=0pt, align=left, font=\fsS] at (6.4600,3.7100) {1,191 stop\\[-1pt]117 wrong (9.8\%)};
\fill[gR!11] (1.5000,3.3000) -- (2.6989,3.3000) -- (6.3600,2.3300) -- (1.5000,2.3300) -- cycle;
\draw[gR!55, line width=0.4pt] (2.6989,3.3000) -- (6.3600,2.3300);
\draw[gR!55, line width=0.4pt] (1.5000,3.3000) -- (1.5000,2.3300);
\node[anchor=north west, inner sep=0pt, align=left, font=\fsS, text=gR!85!black] at (4.6128,3.2300) {Escalated: 40\% have a wrong\\[-1pt]standalone Two-pass prediction};
\node[anchor=north west, inner sep=0pt, align=left, font=\fsL\bfseries, text=gR] at (0,2.3500) {Regrounder\\[-0.5pt]{\fsS\mdseries\color{muted}390 requests}\\[-1.5pt]{\fsS\mdseries\color{muted}up to 2 calls}};
\path[seg, fill=gF] (1.5000,1.9300) rectangle (2.7337,2.3300);
\node[lab, text=white] at (2.1168,2.1300) {99};
\path[seg, fill=tR] (2.7337,1.9300) rectangle (5.4005,2.3300);
\node[lab, text=ink] at (4.0671,2.1300) {214 correct};
\path[seg, fill=wrong] (5.4005,1.9300) rectangle (5.8242,2.3300);
\path[recov] (5.8242,1.9300) rectangle (6.1482,2.3300);
\path[seg, fill=tR] (6.1482,1.9300) rectangle (6.3102,2.3300);
\path[seg, fill=wrong] (6.3102,1.9300) rectangle (6.3475,2.3300);
\path[recov] (6.3475,1.9300) rectangle (6.3600,2.3300);
\node[anchor=north west, inner sep=0pt, align=left, font=\fsS] at (6.4600,2.3400) {291 stop\\[-1pt]64 wrong (22.0\%)};
\draw[gR!70, line width=0.4pt] (6.1482,1.9000) -- (6.1482,1.8500) -- (6.3600,1.8500) -- (6.3600,1.9000);
\node[anchor=north east, inner sep=0pt, font=\fsS, text=gR!85!black] at (6.3600,1.8000) {Context: 17 (13 correct)};
\fill[gF!11] (1.5000,1.9300) -- (2.7337,1.9300) -- (6.3600,0.9600) -- (1.5000,0.9600) -- cycle;
\draw[gF!55, line width=0.4pt] (2.7337,1.9300) -- (6.3600,0.9600);
\draw[gF!55, line width=0.4pt] (1.5000,1.9300) -- (1.5000,0.9600);
\node[anchor=north west, inner sep=0pt, align=left, font=\fsL\bfseries, text=gF] at (0,0.9800) {Finalizer\\[-0.5pt]{\fsS\mdseries\color{muted}99 requests}\\[-1.5pt]{\fsS\mdseries\color{muted}+1 call}};
\path[seg, fill=tF] (1.5000,0.5600) rectangle (3.8564,0.9600);
\node[lab, text=ink] at (2.6782,0.7600) {48 correct};
\path[seg, fill=wrong] (3.8564,0.5600) rectangle (5.5255,0.9600);
\node[lab, text=white] at (4.6909,0.7600) {34 wrong};
\path[recov] (5.5255,0.5600) rectangle (6.3600,0.9600);
\node[lab, text=ink, fill=white, rounded corners=0.5pt] at (5.9427,0.7600) {17};
\node[anchor=north west, inner sep=0pt, align=left, font=\fsS] at (6.4600,0.9700) {48.5\% correct\\[-1pt](Selective: 30.3\%)};
\fill[tB] (0,-0.230) rectangle ++(0.17,0.13); \fill[tR] (0.17,-0.230) rectangle ++(0.17,0.13); \fill[tF] (0.34,-0.230) rectangle ++(0.17,0.13);
\node[anchor=west, font=\fsLeg, inner sep=0pt, xshift=3pt] (lg1) at (0.51,-0.165) {Correct (by stage)};
\fill[wrong] ([xshift=7pt]lg1.east) ++(0,-0.065) rectangle ++(0.17,0.13);
\node[anchor=west, font=\fsLeg, inner sep=0pt] (lg2) at ([xshift={10pt+0.17cm}]lg1.east) {Wrong};
\path[recov] ([xshift=7pt]lg2.east) ++(0,-0.065) rectangle ++(0.17,0.13);
\node[anchor=west, font=\fsLeg, inner sep=0pt] at ([xshift={10pt+0.17cm}]lg2.east) {Wrong after prior correct};
\end{tikzpicture}
\endgroup
\caption{\textbf{How requests move through GroundCascade on SSP} (I32/I8/I32). Each row is rescaled to arriving requests; tinted wedges show escalation to the next stage. The first gate sends \V{gatessp}{acquired} requests on, \V{sspi8}{errEsc}\% of which have a wrong Grounder Detail point, against \V{sspi8}{errStable}\% of the \V{sspi8}{exitsG} that stop. Stop segments are split into correct (tinted by stage), wrong, and wrong although a correct prediction had already been acquired (hatched; \V{erssp}{totalCorrectAvailable} of the \V{erssp}{totalErrors} errors). Supplementary Fig.~\ref{fig:s_flow} shows OSG-R and UIV.}
\label{fig:flow}
\end{figure}

Figure~\ref{fig:flow} follows SSP requests to their exit stage and marks errors made after a correct prediction had already been acquired; supplementary Fig.~\ref{fig:s_flow} does the same for OSG-R and UIV. On SSP, \V{erssp}{gExitErrors} of \V{erssp}{totalErrors} errors exit at the Grounder, and both its predictions are wrong in \V{erssp}{gExitNoneAvailable} of them. The target lies entirely inside the Detail crop for \V{sig}{secin} stable errors and entirely outside it for \V{sig}{secout}, with two partial overlaps. Stable errors therefore include mistakes within an available crop. On SSP, cross-model displacement better ranks incorrect Grounder Detail predictions than self-displacement (AUROC \V{sig}{allcrossmodel} vs.\ \V{sig}{allscreenlinf}), but requires an additional Regrounder Context call.

Some incorrect returns occur even though an earlier acquired prediction is correct. On SSP, this happens for \V{erssp}{agreeExitCorrectAvailable} of \V{erssp}{agreeExitErrors} errors on the Regrounder's Detail-agreement route, \V{erssp}{contextExitCorrectAvailable} of \V{erssp}{contextExitErrors} on its Context-agreement route, and \V{erssp}{fExitCorrectAvailable} of \V{erssp}{fExitErrors} at the Finalizer. On Qwen OSG-R, \V{mixosg}{rExitAvail} Regrounder-exit errors likewise had a correct acquired candidate; on UIV, these counts are \V{mixuiv}{rExitAvail} at $R$ and \V{mixuiv}{fExitAvail} at $F$. The Finalizer recovers \V{erssp}{fNovelCorrect} of \V{erssp}{fNovelCohort} SSP requests despite four incorrect earlier predictions; Fig.~\ref{fig:qualitative} shows two recoveries, one from this cohort.

\section{Conclusion}
\label{sec:conclusion}
\gc uses agreement between two views to decide when to acquire a complementary grounder. Its Selective variant preserves corroborated full-screen predictions alongside zoomed candidates; the Full controller adds a fresh joint-view prediction when disagreements remain. On SSP and OSG-R, \gc improves on the stronger standalone two-pass model by 2.66 and 2.35 points; on UIV, Selective improves by 2.56 points, with no further gain from finalization. Across the three Full allocations, the controller directs complementary predictions toward useful corrections while reducing model invocations by 31--39\% relative to unconditional two-model acquisition.

\label{gc:mainend}
\section*{LLM Usage Statement}
ChatGPT assisted with literature review, paper writing and editing,
recorded-prediction analysis, figures,~and~code.

{\small\bibliographystyle{ieeenat_fullname}\bibliography{references}}

@inproceedings{screenspot,
  author={Kaixin Li and Ziyang Meng and Hongzhan Lin and Ziyang Luo and Yuchen Tian and Jing Ma and Zhiyong Huang and Tat-Seng Chua},
  title={{ScreenSpot-Pro}: {GUI} Grounding for Professional High-Resolution Computer Use}, booktitle={ACM Multimedia}, year={2025}, doi={10.1145/3746027.3755688}}

@inproceedings{osworld,
  author={Tianbao Xie and Jiaqi Deng and Xiaochuan Li and Junlin Yang and Haoyuan Wu and Jixuan Chen and Wenjing Hu and Xinyuan Wang and Yuhui Xu and Zekun Wang and Yiheng Xu and Junli Wang and Doyen Sahoo and Tao Yu and Caiming Xiong},
  title={Scaling Computer-Use Grounding via User Interface Decomposition and Synthesis}, booktitle={NeurIPS Datasets and Benchmarks Track}, year={2025}}

@inproceedings{uivision,
  author={Shravan Nayak and Xiangru Jian and Kevin Qinghong Lin and Juan A. Rodriguez and Montek Kalsi and Nicolas Chapados and M. Tamer {\"O}zsu and Aishwarya Agrawal and David Vazquez and Christopher Pal and Perouz Taslakian and Spandana Gella and Sai Rajeswar},
  title={{UI-Vision}: A Desktop-Centric {GUI} Benchmark for Visual Perception and Interaction}, booktitle={ICML}, year={2025}}

@article{zoomconsistency,
  author={Keon Kim and Krish Chelikavada},
  title={Zoom Consistency: A Free Confidence Signal in Multi-Step Visual Grounding Pipelines}, journal={arXiv:2604.15376}, year={2026}}

@inproceedings{cascade_deferral,
  author={Wittawat Jitkrittum and Neha Gupta and Aditya K. Menon and Harikrishna Narasimhan and Ankit Rawat and Sanjiv Kumar},
  title={When Does Confidence-Based Cascade Deferral Suffice?},
  booktitle={Advances in Neural Information Processing Systems}, volume={36}, year={2023}}

@inproceedings{vtts,
  author={Tiange Luo and Lajanugen Logeswaran and Justin Johnson and Honglak Lee},
  title={Visual Test-Time Scaling for {GUI} Agent Grounding}, booktitle={ICCV}, year={2025}}

@article{zoomclick,
  author={Zhiyuan Jiang and Shenghao Xie and Wenyi Li and Wenqiang Zu and Peihang Li and Jiahao Qiu and Siqi Pei and Lei Ma and Tiejun Huang and Mengdi Wang and Shilong Liu},
  title={Zoom in, Click out: Unlocking and Evaluating the Potential of Zooming for {GUI} Grounding}, journal={arXiv:2512.05941}, year={2025}}

@article{uizoomer,
  author={Fei Tang and Bofan Chen and Zhengxi Lu and Tongbo Chen and Songqin Nong and Tao Jiang and Wenhao Xu and Weiming Lu and Jun Xiao and Yueting Zhuang and Yongliang Shen},
  title={{UI-Zoomer}: Uncertainty-Driven Adaptive Zoom-In for {GUI} Grounding}, journal={arXiv:2604.14113}, year={2026}}

@inproceedings{drsgui,
  author={Yichao Liu and Huawen Shen and Liu Yu and Shiyu Liu and Zeyu Chen and Yu Zhou},
  title={{DRS-GUI}: Dynamic Region Search for Training-Free {GUI} Grounding}, booktitle={CVPR}, year={2026}}

@article{rankground,
  author={Liyang Fan and Xinping Bi and Yitai Li and Shuaimin Li and Hui Li and Min Yang},
  title={{RankGround}: Efficient High-Resolution {GUI} Grounding via Lightweight Reranker-Guided Crop Selection}, journal={arXiv:2609.18690}, year={2026}}

@inproceedings{mvp,
  author={Yunzhu Zhang and Zeyu Pan and Zhengwen Zeng and Shuheng Shen and Changhua Meng and Linchao Zhu},
  title={{MVP}: Multiple View Prediction Improves {GUI} Grounding}, booktitle={CVPR}, year={2026}}

@article{safeground,
  author={Qingni Wang and Yue Fan and Xin Eric Wang},
  title={{SafeGround}: Know When to Trust {GUI} Grounding Models via Uncertainty Calibration}, journal={arXiv:2602.02419}, year={2026}}

@inproceedings{argus,
  author={Divake Kumar and Sina Tayebati and Devashri Naik and Amanda Sofie Rios and Nilesh Ahuja and Omesh Tickoo and Ranganath Krishnan and Amit Ranjan Trivedi},
  title={{Argus}: A Cross-Regime Benchmark for the Transferability of Uncertainty Quantification in Computer-Use Agents}, booktitle={NeurIPS Evaluations and Datasets Track}, year={2026}}

@article{lookagain,
  author={Renshan Zhang and Haoyang Meng and Yixiao He and Rui Shao and April Hua Liu and Liqiang Nie},
  title={{LookAgain}: Closed-Loop {GUI} Grounding with Visually Grounded Reflection}, journal={arXiv:2608.09723}, year={2026}}

@article{reprefill,
  author={Jiaping Lin and Fei Shen and Junzhe Li and Ping Nie and Fei Yu and Ming Li and Haizhou Li},
  title={What Happens Before Decoding? {P}refill Determines {GUI} Grounding in {VLMs}}, journal={arXiv:2605.12549}, year={2026}}

@article{maiui,
  author={Hanzhang Zhou and Xu Zhang and Panrong Tong and Jianan Zhang and Liangyu Chen and Quyu Kong and Chenglin Cai and Chen Liu and Yue Wang and Jingren Zhou and Steven Hoi},
  title={{MAI-UI} Technical Report: Real-World Centric Foundation {GUI} Agents}, journal={arXiv:2512.22047}, year={2025}}

@misc{indeedui,
  author={{Intelligence Indeed}},
  title={{Indeed-UI}: High-Resolution {GUI} Grounding Model}, howpublished={\url{https://github.com/intelligence-indeed/Indeed-UI}}, year={2026}}

@misc{kvground,
  author={{Kingsware and Vocaela AI}},
  title={{KV-Ground}: Small {GUI} Grounding Models for High-Resolution Images}, howpublished={\url{https://github.com/vocaela/kv-ground}}, year={2026}}

@misc{qwen36card,
  author={{Qwen Team}},
  title={{Qwen3.6-27B}: Model Card},
  howpublished={\url{https://huggingface.co/Qwen/Qwen3.6-27B}}, year={2026}}

@misc{qwen38card,
  author={{Qwen Team}},
  title={{Qwen3.8-27B}: Model Card},
  howpublished={\url{https://huggingface.co/Qwen/Qwen3.8-27B}}, year={2026}}

@inproceedings{groundnext,
  author={Aarash Feizi and Shravan Nayak and Xiangru Jian and Kevin Qinghong Lin and Kaixin Li and Rabiul Awal and Xing Han L{\`u} and Johan Obando-Ceron and Juan A. Rodriguez and Nicolas Chapados and David Vazquez and Adriana Romero-Soriano and Reihaneh Rabbany and Perouz Taslakian and Christopher Pal and Spandana Gella and Sai Rajeswar},
  title={Grounding Computer Use Agents on Human Demonstrations}, booktitle={ICLR}, year={2026}}

@inproceedings{violajones,
  author={Paul Viola and Michael Jones},
  title={Rapid Object Detection using a Boosted Cascade of Simple Features}, booktitle={CVPR}, year={2001}}

@inproceedings{wisdom,
  author={Xiaofang Wang and Dan Kondratyuk and Eric Christiansen and Kris M. Kitani and Yair Alon and Elad Eban},
  title={Wisdom of Committees: An Overlooked Approach to Faster and More Accurate Models}, booktitle={ICLR}, year={2022}}

@article{frugalgpt,
  author={Lingjiao Chen and Matei Zaharia and James Zou},
  title={{FrugalGPT}: How to Use Large Language Models While Reducing Cost and Improving Performance}, journal={Transactions on Machine Learning Research}, year={2024}}

@article{venus2,
  title   = {{UI-Venus-2} Technical Report},
  author  = {{Venus Team} and Zhuohan Cai and Haoxing Chen and Jiaxuan Chen and Weizhi Chen and Changlong Gao and Zhangxuan Gu and Yuan Guo and Yusong Hu and Jianrong Jiang and Jianguo Li and Runze Li and Jinzhen Lin and Zhenyu Ma and Changhua Meng and Han Peng and Xinyu Qiu and Shuheng Shen and Zhongyi Shui and Weiqiang Wang and Ming Wen and Zhuoer Xu and Hang Yan and Kaiwen Yang and Ruilin Yao and Nanjun Yu and Zhengwen Zeng and Lianrui Zhang and Yunzhu Zhang and Zhe Zhao and Beitong Zhou},
  journal = {arXiv:2609.00028},
  year    = {2026}
}

@article{qwenuiagent,
  title   = {{Qwen-UI-Agent} Technical Report: Toward Next-Generation Real-World Centric Foundation {GUI} Agents},
  author  = {Hanzhang Zhou and Panrong Tong and Xu Zhang and Quyu Kong and Chenglin Cai and Tianyu Xia and Gongjie Zhang and Jianan Zhang and Long Li and Long Chen and Lei Wang and Gaole Dai and Pengxiang Li and Liangyu Chen and Yue Wang and Steven Hoi},
  journal = {arXiv:2607.28227},
  year    = {2026}
}

@article{autofocus,
  title   = {{AutoFocus}: Uncertainty-Aware Active Visual Search for {GUI} Grounding},
  author  = {Ruilin Yao and Shegnwu Xiong and Tianyu Zou and Shili Xiong and Yi Rong},
  journal = {arXiv:2605.02630},
  year    = {2026}
}

@article{avr,
  title   = {Adaptive Vision-Language Model Routing for Computer Use Agents},
  author  = {Xunzhuo Liu and Bowei He and Xue Liu and Andy Luo and Haichen Zhang and Huamin Chen},
  journal = {arXiv:2603.12823},
  year    = {2026}
}

@inproceedings{copc,
  title     = {Measure Twice, Click Once: Co-evolving Proposer and Visual Critic via Reinforcement Learning for {GUI} Grounding},
  author    = {Wenkai Wang and Xiyun Li and Hongcan Guo and Wenhao Yu and Tianqing Fang and Haitao Mi and Dong Yu and Shengyu Zhang},
  booktitle = {ACL},
  year      = {2026}
}

@inproceedings{motcascade,
  author={Murong Yue and Jie Zhao and Min Zhang and Liang Du and Ziyu Yao},
  title={Large Language Model Cascades with Mixture of Thought Representations for Cost-Efficient Reasoning}, booktitle={ICLR}, year={2024}}

@article{pace,
  author={Yunxiang Li and Xixin Wu and Helen M. Meng},
  title={How Many Pixels Is a Digit Worth? Place-Aware Coordinate Entropy for {GUI} Agent Confidence Estimation},
  journal={arXiv:2609.24277}, note={Accepted to EMNLP 2026}, year={2026}}

@article{chainofground,
  author={Aiden Yiliu Li and Bizhi Yu and Daoan Lei and Tianhe Ren and Shilong Liu},
  title={{Chain-of-Ground}: Improving {GUI} Grounding via Iterative Reasoning and Reference Feedback}, journal={arXiv:2512.01979}, year={2025}}

@article{gms,
  author={Zhecheng Li and Guoxian Song and Yiwei Wang and Zhen Xiong and Junsong Yuan and Yujun Cai},
  title={Generalist Scanner Meets Specialist Locator: A Synergistic Coarse-to-Fine Framework for Robust {GUI} Grounding}, journal={arXiv:2509.24133}, year={2025}}

@article{mhvf,
  author={Minghua Chen and Xingming Liao and Weiqiang Zhong and Zhuowei Wang},
  title={{MHVF}: Training-Free {GUI} Grounding via Verification-Guided Multi-Hypothesis Search}, journal={arXiv:2610.17792}, year={2026}}

@misc{screenspot_leaderboard,
  author={{ScreenSpot-Pro Benchmark Maintainers}},
  title={{ScreenSpot-Pro Leaderboard}},
  howpublished={\url{https://gui-agent.github.io/grounding-leaderboard/}},
  note={Accessed 5 November 2026}, year={2026}}

@misc{hunmin_cua,
  author={{MNC AI}},
  title={{Hunmin-CUA}: Post-Training a 397B {MoE} for Computer Use},
  howpublished={\url{https://huggingface.co/blog/mncai/hunmin-cua-post-training-a-397b-moe-for-computer-u}},
  year={2026}}

@article{leash,
  author={Qingni Wang and Changdae Oh and Yuzhe Yang and Sharon Li and Jiachen Yang and Ang Li and Xin Eric Wang},
  title={{LEASH}: Uncertainty-Aware Weak-to-Strong Escalation and Oversight for Computer-Use Agents},
  journal={arXiv:2610.22280}, year={2026}}

@inproceedings{bami,
  author={Borui Zhang and Bo Zhang and Bo Wang and Wenzhao Zheng and Yuhao Cheng and Liang Tang and Yiqiang Yan and Jie Zhou and Jiwen Lu},
  title={{BAMI}: Training-Free Bias Mitigation in {GUI} Grounding},
  booktitle={Proceedings of the IEEE/CVF Conference on Computer Vision and Pattern Recognition},
  pages={34596--34605}, year={2026}}

@inproceedings{fastfocus,
  author={Zhenhua Yang and Xiachong Feng and Weihong Zhong and Lunjun Liu and Xiaocheng Feng and Bing Qin},
  title={Focus Like a Human: Efficient {GUI} Grounding via Coarse-to-Fine Visual Attention and Parallel Verification},
  booktitle={Proceedings of the Thirty-Fifth International Joint Conference on Artificial Intelligence},
  pages={428--436}, year={2026}, doi={10.24963/ijcai.2026/49}}

@article{guilens,
  author={Zichuan Fu and Shirong Wang and Wenlin Zhang and Guojing Li and Yimin Deng and Jingtong Gao and Junjia Qi and Hanyu Yan and Yefeng Zheng and Xiaopeng Li and Wanyu Wang and Xian Wu and Xiangyu Zhao},
  title={{GUI-Lens}: Coarse-to-Fine Cropping for {GUI} Grounding with General-Purpose {VLMs}},
  journal={arXiv:2608.03270}, year={2026}}

@article{guiowl15,
  title   = {{Mobile-Agent-v3.5}: Multi-platform Fundamental {GUI} Agents},
  author  = {Haiyang Xu and Xi Zhang and Haowei Liu and Junyang Wang and Zhaozai Zhu and Shengjie Zhou and Xuhao Hu and Feiyu Gao and Junjie Cao and Zihua Wang and Zhiyuan Chen and Jitong Liao and Qi Zheng and Jiahui Zeng and Ze Xu and Shuai Bai and Junyang Lin and Jingren Zhou and Ming Yan},
  journal = {arXiv:2602.16855},
  year    = {2026}
}

@article{venus15,
  title   = {{UI-Venus-1.5} Technical Report},
  author  = {{Venus Team} and Changlong Gao and Zhangxuan Gu and Yulin Liu and Xinyu Qiu and Shuheng Shen and Yue Wen and Tianyu Xia and Zhenyu Xu and Zhengwen Zeng and Beitong Zhou and Xingran Zhou and Weizhi Chen and Sunhao Dai and Jingya Dou and Yichen Gong and Yuan Guo and Zhenlin Guo and Feng Li and Qian Li and Jinzhen Lin and Yuqi Zhou and Linchao Zhu and Liang Chen and Zhenyu Guo and Changhua Meng and Weiqiang Wang},
  journal = {arXiv:2602.09082},
  year    = {2026}
}

@misc{holo2,
  author       = {{H Company}},
  title        = {{H Company}'s New {Holo2} Model Takes the Lead in {UI} Localization},
  howpublished = {\url{https://huggingface.co/blog/Hcompany/introducing-holo2-235b-a22b}},
  note         = {3 February 2026},
  year         = {2026}
}

@misc{gpt6astra,
  author       = {{OpenAI}},
  title        = {{GPT-6 Astra}: A New Generation of Intelligence},
  howpublished = {\url{https://openai.com/index/gpt-6-astra/}},
  note         = {3 September 2026},
  year         = {2026}
}
\clearpage
\setcounter{section}{0}\setcounter{equation}{0}\setcounter{table}{0}\setcounter{figure}{0}\setcounter{algorithm}{0}
\renewcommand{\thesection}{S\arabic{section}}
\renewcommand{\theequation}{S\arabic{equation}}
\renewcommand{\thetable}{S\arabic{table}}
\renewcommand{\thefigure}{S\arabic{figure}}
\renewcommand{\thealgorithm}{S\arabic{algorithm}}
\renewcommand{\theHsection}{supp.\arabic{section}}
\renewcommand{\theHequation}{supp.\arabic{equation}}
\renewcommand{\theHtable}{supp.\arabic{table}}
\renewcommand{\theHfigure}{supp.\arabic{figure}}
\providecommand{\theHalgorithm}{\arabic{algorithm}}
\renewcommand{\theHalgorithm}{supp.\arabic{algorithm}}
\maketitlesupplementary
\section{Controller and inference}
\label{sec:s_impl}
Algorithm~\ref{alg:s_gc} gives the complete routing procedure.
Selective executes the algorithm through both Stage~2 agreement tests and returns the selected Detail point if neither passes; it omits only the Finalizer.
\begin{algorithm}[b]
\caption{\textbf{GroundCascade inference.} $I$ is the screenshot, $q$ the instruction, and $G,R,F$ the Grounder, Regrounder, and Finalizer. \textsc{TwoPass} skips Detail after invalid Context; \textsc{FirstValid} returns the first valid point or invalid.}
\label{alg:s_gc}
\begin{algorithmic}[1]
\Statex\hspace*{-\leftmargin}\textsc{GroundCascade}$(I,q,G,R,F)$
\State $(p_{G,C},p_{G,D})\gets\textsc{TwoPass}(G,I,q)$
\State $s_G\gets d(p_{G,C},p_{G,D})$
\If{$s_G\le\tau_s$}
  \State \Return $p_{G,D}$
\EndIf
\State $(p_{R,C},p_{R,D})\gets\textsc{TwoPass}(R,I,q)$
\State $s_R\gets d(p_{R,C},p_{R,D})$
\State $m^\star\gets G$ if $s_G\le s_R$, else $R$
\State $p_D^\star\gets\textsc{FirstValid}(p_{m^\star,D},p_{G,D},p_{R,D})$
\If{$d(p_{G,D},p_{R,D})\le\tau_D$}
  \State \Return $p_D^\star$
\EndIf
\If{$d(p_{G,C},p_{R,C})\le\tau_C$}
  \State \Return $p_{m^\star,C}$
\EndIf
\State $J\gets\textsc{JointCrop}(I,p_{G,C},p_{G,D},p_{R,C},p_{R,D})$
\State $p_{F,J}\gets F(J,q)$ mapped to $I$
\State \Return $\textsc{FirstValid}(p_{F,J},p_D^\star,p_{G,C},p_{R,C})$
\end{algorithmic}
\end{algorithm}

\smallhead{Inference settings.}
The Indeed/KV experiments used BF16, FlashAttention~2, deterministic decoding with at most 512 new tokens, and an image budget of approximately $10^8$ pixels. The Qwen runs use the same click-coordinate interface, disabled thinking, temperature zero, and their separately pinned BF16 vLLM checkpoints. Each call receives one view and the original instruction; GPU inference never receives target annotations.

\smallhead{Views and coordinates.}
For a valid Context point $(x,y)$ in integer screenshot coordinates, let $a_x=\lfloor W/4\rfloor$ and $a_y=\lfloor H/4\rfloor$. The Detail crop $[l,r)\times[t,b)$ has bounds
\begin{align}
 l&=\max(0,x-a_x), & r&=\min(W,x+a_x),\notag\\
 t&=\max(0,y-a_y), & b&=\min(H,y+a_y).
 \label{eq:s_native}
\end{align}
Both cropped views are unmarked and resized to $W\times H$ by bicubic interpolation. Joint crops retain their rounded dimensions when shifted inside the screenshot.

For a prediction $(u,v)$ on the 0--1000 grid, round to view pixels $(\hat u,\hat v)=(\mathrm{round}(Wu/1000),\mathrm{round}(Hv/1000))$, then map back:
\begin{equation}
 p=\bigl(\mathrm{round}[l+\tfrac{\hat u}{W}(r-l)],\mathrm{round}[t+\tfrac{\hat v}{H}(b-t)]\bigr).
 \label{eq:s_map}
\end{equation}
Rounding uses ties to even; Context spans $[0,W)\times[0,H)$. The parser accepts only the first tool call, which must be non-refusal and contain two finite, nonnegative coordinates. Rounded view and mapped screenshot pixels must lie in bounds; invalid outputs are not clipped.

\newcommand{\FinalizerSuppTable}{%
\begin{table}[H]\centering\footnotesize\setlength{\tabcolsep}{2.5pt}
\begin{tabular*}{\columnwidth}{@{\extracolsep{\fill}}llrrrr@{}}\toprule
Benchmark & $G/R/F$ & $F$ req. & Rep. & Reg. & $\Delta$ pp\\\midrule
SSP & I32/I8/I32 & \V{erssp}{fRequests} & \V{erssp}{fRepairs} & \V{erssp}{fRegressions} & $+$1.14\\
OSG-R & Q36/Q38/Q38 & \V{mixosg}{fReq} & \V{mixosg}{fRep} & \V{mixosg}{fReg} & \V{mixosg}{stage3pp}\\
UIV & Q36/Q38/Q38 & \V{mixuiv}{fReq} & \V{mixuiv}{fRep} & \V{mixuiv}{fReg} & \V{mixuiv}{stage3pp}\\
\bottomrule\end{tabular*}
\caption{\textbf{Finalizer effect relative to Selective.} Repairs and regressions count corrected and spoiled Stage~2 answers. $\Delta$ is the whole-benchmark accuracy change in percentage points (category-balanced for UIV).}
\label{tab:s_finalizer}
\end{table}}

\section{Evaluation details}
\label{sec:s_metrics}
\smallhead{Scoring.}
Target boxes include their boundaries. OSG-R consists of 470 rectangles and 40 polygons; polygons use the benchmark's ray-crossing test. UIV contains 1,772 Basic, 1,772 Functional, and 1,935 Spatial queries across 1,181 screenshot files representing 1,156 distinct images. Its overall score equally averages category accuracies.

\smallhead{Published OSG-R bounds.}
The original benchmark has 564 requests, including 510 visible-target cases. If a published one-decimal percentage $a$ is single-run micro-accuracy on all 564, its visible-target accuracy is at most $(a+0.05)\cdot564/510$; Fig.~1 rounds this ceiling upward to 0.1 point. The four reported scores 79.1, 78.5, 78.5, and 76.4\% correspond to 446, 443, 443, and 431 correct of 564, respectively, but none can arise by rounding integer correct counts out of 510. This supports, without proving, the 564-item micro-accuracy interpretation: averaging runs or categories could also yield those percentages. Refusal outcomes are unavailable. The starred, hatched $\leq$ bars are therefore \emph{conditional bounds}, not observed subset scores or matched evaluations.

\smallhead{MVP interface.}
Released MVP uses I32 on all 1,581 SSP requests with its own prompt, parser, attention-guided views and four-call interface; its image cap is $4096\times2160$ with 32 output tokens.

\smallhead{Controls and uncertainty.}
The standalone Two-pass baseline returns valid Detail, otherwise valid Context; invalid Context skips Detail. Call savings compare Full's executed invocations with acquiring both models' Two-pass predictions on every request, including the same invalid-Context skip. This unconditional control returns the valid Detail with smaller Context--Detail displacement (ties favor the Grounder), falling back to the other valid Detail; it omits Context return and finalization.

Paired 95\% accuracy intervals use 10,000 bootstrap resamples (seed 20271001): by request for SSP and by RGB-identical image for OSG-R and UIV. All UIV queries sharing an image are sampled together across categories.

\smallhead{Second-opinion budget control.}
We compare each flagship acquisition gate against 50,000 random allocations using recorded predictions and the same Stage~2 policy. Random requests are sampled without replacement within recorded Regrounder-call-cost strata, additionally preserving UIV category budgets; each stratum receives the same number of acquisitions as the gate. This is a retrospective, executed-cost-matched control.

\begin{table}[H]
\centering\footnotesize\setlength{\tabcolsep}{2.4pt}
\begin{tabular*}{\columnwidth}{@{\extracolsep{\fill}}lrrrr@{}}\toprule
Benchmark & $R$ requests & Gate gain & Random mean & Random 95\% \\\midrule
SSP & \V{gatessp}{acquired} & \V{gatessp}{actualPp} & \V{gatessp}{uniformMeanPp} & [\V{gatessp}{uniformLowPp},\,\V{gatessp}{uniformHighPp}]\\
OSG-R & \V{mixosg}{budgetAcquired} & \V{mixosg}{budgetActualPp} & \V{mixosg}{budgetUniformMeanPp} & [\V{mixosg}{budgetUniformLowPp},\,\V{mixosg}{budgetUniformHighPp}]\\
UIV & \V{mixuiv}{budgetAcquired} & \V{mixuiv}{budgetActualPp} & \V{mixuiv}{budgetUniformMeanPp} & [\V{mixuiv}{budgetUniformLowPp},\,\V{mixuiv}{budgetUniformHighPp}]\\
\bottomrule\end{tabular*}
\caption{\textbf{Selective acquisition at fixed cost.} Gains over standalone Two-pass (percentage points). Random columns show the mean and central 95\% range across stratified random allocations; UIV uses category-balanced accuracy. Each row uses its flagship model allocation.}
\label{tab:s_helper_budget}
\end{table}

\section{Cross-model and category analysis}
\label{sec:s_robust}
\smallhead{Model roles.}
On SSP, I32/I8/I32 and I32/K8/I32 reach \V{sspi8}{gc}\% and \V{sspk8}{gc}\%, versus \V{ord}{i8acc}\% and \V{ord}{k8acc}\% with $G$ and $R$ swapped. The Q38 Regrounder also trails the Q36 Grounder in standalone Two-pass accuracy on OSG-R (\V{mixosg}{rdetail}\% vs. \V{mixosg}{detail}\%) and UIV (\V{mixuiv}{rdetail}\% vs. \V{mixuiv}{detail}\%), yet contributes complementary predictions.

\smallhead{UIV categories.}
Selective outperforms Q36 Two-pass in all UIV categories (Tab.~\ref{tab:s_uiv}). Full further improves Basic and Functional over Selective, but loses 11 net correct Spatial predictions across \V{mixuiv}{spatial_fReq} Finalizer calls; its category-balanced change from Selective is \V{mixuiv}{stage3pp} points.

\begin{table}[H]
\centering\footnotesize\setlength{\tabcolsep}{2.5pt}
\begin{tabular*}{\columnwidth}{@{\extracolsep{\fill}}lrrrrr@{}}\toprule
UIV category & Context & Two-pass & Selective & Full & $F$ calls\\\midrule
Basic & \V{mixuiv}{basic_ctx} & \V{mixuiv}{basic_detail} & \V{mixuiv}{basic_sel} & \V{mixuiv}{basic_full} & \V{mixuiv}{basic_fReq}\\
Functional & \V{mixuiv}{functional_ctx} & \V{mixuiv}{functional_detail} & \V{mixuiv}{functional_sel} & \V{mixuiv}{functional_full} & \V{mixuiv}{functional_fReq}\\
Spatial & \V{mixuiv}{spatial_ctx} & \V{mixuiv}{spatial_detail} & \V{mixuiv}{spatial_sel} & \V{mixuiv}{spatial_full} & \V{mixuiv}{spatial_fReq}\\\midrule
Mean & \V{mixuiv}{ctx} & \V{mixuiv}{detail} & \V{mixuiv}{sel} & \V{mixuiv}{gc} & \V{mixuiv}{fReq}\\
\bottomrule\end{tabular*}
\caption{\textbf{Q36/Q38/Q38 on UI-Vision.} Category accuracy (\%) and Finalizer invocations. Context and Two-pass use the standalone Q36 Grounder; Mean is the official equal-weight category score.}
\label{tab:s_uiv}
\end{table}

\smallhead{One-pass interface check.}
The UI-Venus-2 report lists 58.3\% on UIV for its reproduced Qwen3.6-27B baseline, without specifying the aggregation. With its released coordinate interface and our pinned Q36 checkpoint, a separate run scores \V{venus}{correct}/\V{venus}{n} (\V{venus}{pooled}\% pooled; \V{venus}{macro}\% category-balanced), versus \V{mixuiv}{ctx}\% balanced for our structured-click Context. Prompts and serving conditions differ, so this comparison does not isolate an interface effect.

\section{Finalizer and joint view}
\label{sec:s_geom}
\smallhead{Matched-budget allocation.}
On SSP, \V{ivssp}{FSelectedN} gated Finalizer calls add \V{ivssp}{FSelectedNet} correct answers (Tab.~\ref{tab:s_finalizer}); uniform allocation among \V{erssp}{gGateDeferred} eligible requests yields \V{ivssp}{FUniformMeanOne} in expectation.
\FinalizerSuppTable

\smallhead{Cross-model effect.}
On Qwen UIV, \V{mixuiv}{fRep} Finalizer repairs and \V{mixuiv}{fReg} regressions yield five fewer correct answers: six net repairs in Basic and Functional, offset by eleven Spatial losses (Tab.~\ref{tab:s_uiv}).

\section{Error and recovery diagnostics}
\label{sec:s_signals}
Table~\ref{tab:s_error_recovery} reports gate detection and recovery for all three flagship allocations.
\ErrorRecoverySuppTable
\smallhead{Error-ranking AUROC.}
We rank standalone Two-pass errors by $d(p_{G,C},p_{G,D})$ (within-model) or $d(p_{G,D},p_{R,C})$ (cross-model, cross-view). Their AUROCs on escalations are \V{sig}{sspEscSelfAUROC}/\V{sig}{sspEscCrossAUROC} for SSP, \V{mixosg}{selfAUROC}/\V{mixosg}{crossAUROC} for OSG-R, and \V{mixuiv}{selfAUROC}/\V{mixuiv}{crossAUROC} for UIV. Across all SSP requests, they reach \V{sig}{allscreenlinf}/\V{sig}{allcrossmodel} (paired gain CI: \V{sig}{cixs}); cross-model ranking requires an $R$ call.

\smallhead{Context preservation with Qwen.}
On UIV, the strict Context-agreement exit is taken on \V{mixuiv}{ctxReq} requests, correcting \V{mixuiv}{ctxRep} and regressing \V{mixuiv}{ctxReg} Detail predictions (net \V{mixuiv}{ctxNet}). On OSG-R, the corresponding \V{mixosg}{ctxReq} exits have net effect \V{mixosg}{ctxNet}. This isolates Context preservation from acquisition.

\section{Additional request-flow diagnostics}
\label{sec:s_figs}
\begin{figure*}[t]
\centering
\begingroup
\colorlet{gB}{gcG}\colorlet{gBk}{gcGk}\colorlet{gR}{gcR}\colorlet{gF}{gcF}
\colorlet{tB}{gcGt}\colorlet{tR}{gcRt}\colorlet{tF}{gcFt}
\definecolor{wrong}{HTML}{4B5563}\definecolor{ink}{HTML}{2B2F36}\definecolor{muted}{HTML}{6B7280}
\newcommand{\fsS}{\scriptsize}
\newcommand{\fsL}{\footnotesize}
\newcommand{\fsLeg}{\footnotesize}
\tikzset{
  seg/.style={draw=white, line width=0.35pt},
  recov/.style={seg, fill=white, postaction={pattern=north east lines, pattern color=wrong!55}},
  lab/.style={font=\fsS, inner sep=0.6pt},
}
\begin{tikzpicture}
\node[anchor=north west, inner sep=0pt, align=left, font=\fsL\bfseries, text=gBk] at (0,3.7200) {Grounder\\[-0.5pt]{\fsS\mdseries\color{muted}510 requests}\\[-1.5pt]{\fsS\mdseries\color{muted}up to 2 calls}};
\path[seg, fill=gR] (1.5000,3.3000) rectangle (2.4244,3.7000);
\node[lab, text=white] at (1.9622,3.5000) {97};
\path[seg, fill=tB] (2.4244,3.3000) rectangle (6.0836,3.7000);
\node[lab, text=ink] at (4.2540,3.5000) {384 correct (stable)};
\path[seg, fill=wrong] (6.0836,3.3000) rectangle (6.3505,3.7000);
\path[recov] (6.3505,3.3000) rectangle (6.3600,3.7000);
\node[anchor=north west, inner sep=0pt, align=left, font=\fsS] at (6.4600,3.7100) {413 stop\\[-1pt]29 wrong (7.0\%)};
\fill[gR!11] (1.5000,3.3000) -- (2.4244,3.3000) -- (6.3600,2.3300) -- (1.5000,2.3300) -- cycle;
\draw[gR!55, line width=0.4pt] (2.4244,3.3000) -- (6.3600,2.3300);
\draw[gR!55, line width=0.4pt] (1.5000,3.3000) -- (1.5000,2.3300);
\node[anchor=north west, inner sep=0pt, align=left, font=\fsS, text=gR!85!black] at (4.4713,3.2300) {Escalated: 43\% have a wrong\\[-1pt]standalone Two-pass prediction};
\node[anchor=north west, inner sep=0pt, align=left, font=\fsL\bfseries, text=gR] at (0,2.3500) {Regrounder\\[-0.5pt]{\fsS\mdseries\color{muted}97 requests}\\[-1.5pt]{\fsS\mdseries\color{muted}up to 2 calls}};
\path[seg, fill=gF] (1.5000,1.9300) rectangle (2.5522,2.3300);
\node[lab, text=white] at (2.0261,2.1300) {21};
\path[seg, fill=tR] (2.5522,1.9300) rectangle (5.3579,2.3300);
\node[lab, text=ink] at (3.9551,2.1300) {56 correct (agreement)};
\path[seg, fill=wrong] (5.3579,1.9300) rectangle (5.6085,2.3300);
\path[recov] (5.6085,1.9300) rectangle (6.3600,2.3300);
\node[lab, text=ink, fill=white, rounded corners=0.5pt] at (5.9842,2.1300) {15};
\node[anchor=north west, inner sep=0pt, align=left, font=\fsS] at (6.4600,2.3400) {76 stop\\[-1pt]20 wrong (26.3\%)};
\fill[gF!11] (1.5000,1.9300) -- (2.5522,1.9300) -- (6.3600,0.9600) -- (1.5000,0.9600) -- cycle;
\draw[gF!55, line width=0.4pt] (2.5522,1.9300) -- (6.3600,0.9600);
\draw[gF!55, line width=0.4pt] (1.5000,1.9300) -- (1.5000,0.9600);
\node[anchor=north west, inner sep=0pt, align=left, font=\fsL\bfseries, text=gF] at (0,0.9800) {Finalizer\\[-0.5pt]{\fsS\mdseries\color{muted}21 requests}\\[-1.5pt]{\fsS\mdseries\color{muted}+1 call}};
\path[seg, fill=tF] (1.5000,0.5600) rectangle (4.0457,0.9600);
\node[lab, text=ink] at (2.7729,0.7600) {11 correct};
\path[seg, fill=wrong] (4.0457,0.5600) rectangle (5.6657,0.9600);
\node[lab, text=white] at (4.8557,0.7600) {7 wrong};
\path[recov] (5.6657,0.5600) rectangle (6.3600,0.9600);
\node[lab, text=ink, fill=white, rounded corners=0.5pt] at (6.0129,0.7600) {3};
\node[anchor=north west, inner sep=0pt, align=left, font=\fsS] at (6.4600,0.9700) {52.4\% correct\\[-1pt](Selective: 42.9\%)};
\path (0,0.30) -- (0,3.84);

\end{tikzpicture}
\endgroup\hfill
\begingroup
\colorlet{gB}{gcG}\colorlet{gBk}{gcGk}\colorlet{gR}{gcR}\colorlet{gF}{gcF}
\colorlet{tB}{gcGt}\colorlet{tR}{gcRt}\colorlet{tF}{gcFt}
\definecolor{wrong}{HTML}{4B5563}\definecolor{ink}{HTML}{2B2F36}\definecolor{muted}{HTML}{6B7280}
\newcommand{\fsS}{\scriptsize}
\newcommand{\fsL}{\footnotesize}
\newcommand{\fsLeg}{\footnotesize}
\tikzset{
  seg/.style={draw=white, line width=0.35pt},
  recov/.style={seg, fill=white, postaction={pattern=north east lines, pattern color=wrong!55}},
  lab/.style={font=\fsS, inner sep=0.6pt},
}
\begin{tikzpicture}
\node[anchor=north west, inner sep=0pt, align=left, font=\fsL\bfseries, text=gBk] at (0,3.7200) {Grounder\\[-0.5pt]{\fsS\mdseries\color{muted}5,479 requests}\\[-1.5pt]{\fsS\mdseries\color{muted}up to 2 calls}};
\path[seg, fill=gR] (1.5000,3.3000) rectangle (3.0328,3.7000);
\node[lab, text=white] at (2.2664,3.5000) {1,728};
\path[seg, fill=tB] (3.0328,3.3000) rectangle (5.8065,3.7000);
\node[lab, text=ink] at (4.4196,3.5000) {3,127 correct (stable)};
\path[seg, fill=wrong] (5.8065,3.3000) rectangle (6.3538,3.7000);
\path[recov] (6.3538,3.3000) rectangle (6.3600,3.7000);
\node[anchor=north west, inner sep=0pt, align=left, font=\fsS] at (6.4600,3.7100) {3,751 stop\\[-1pt]624 wrong (16.6\%)};
\fill[gR!11] (1.5000,3.3000) -- (3.0328,3.3000) -- (6.3600,2.3300) -- (1.5000,2.3300) -- cycle;
\draw[gR!55, line width=0.4pt] (3.0328,3.3000) -- (6.3600,2.3300);
\draw[gR!55, line width=0.4pt] (1.5000,3.3000) -- (1.5000,2.3300);
\node[anchor=north west, inner sep=0pt, align=left, font=\fsS, text=gR!85!black] at (4.7849,3.2300) {Escalated: 60\% have a wrong\\[-1pt]standalone Two-pass prediction};
\node[anchor=north west, inner sep=0pt, align=left, font=\fsL\bfseries, text=gR] at (0,2.3500) {Regrounder\\[-0.5pt]{\fsS\mdseries\color{muted}1,728 requests}\\[-1.5pt]{\fsS\mdseries\color{muted}up to 2 calls}};
\path[seg, fill=gF] (1.5000,1.9300) rectangle (3.1734,2.3300);
\node[lab, text=white] at (2.3367,2.1300) {595 disagree};
\path[seg, fill=tR] (3.1734,1.9300) rectangle (4.9706,2.3300);
\node[lab, text=ink] at (4.0720,2.1300) {639 correct};
\path[seg, fill=wrong] (4.9706,1.9300) rectangle (5.5809,2.3300);
\node[lab, text=white] at (5.2758,2.1300) {217};
\path[recov] (5.5809,1.9300) rectangle (6.3600,2.3300);
\node[lab, text=ink, fill=white, rounded corners=0.5pt] at (5.9705,2.1300) {277};
\node[anchor=north west, inner sep=0pt, align=left, font=\fsS] at (6.4600,2.3400) {1,133 stop\\[-1pt]494 wrong (43.6\%)};
\fill[gF!11] (1.5000,1.9300) -- (3.1734,1.9300) -- (6.3600,0.9600) -- (1.5000,0.9600) -- cycle;
\draw[gF!55, line width=0.4pt] (3.1734,1.9300) -- (6.3600,0.9600);
\draw[gF!55, line width=0.4pt] (1.5000,1.9300) -- (1.5000,0.9600);
\node[anchor=north west, inner sep=0pt, align=left, font=\fsL\bfseries, text=gF] at (0,0.9800) {Finalizer\\[-0.5pt]{\fsS\mdseries\color{muted}595 requests}\\[-1.5pt]{\fsS\mdseries\color{muted}+1 call}};
\path[seg, fill=tF] (1.5000,0.5600) rectangle (3.0111,0.9600);
\node[lab, text=ink] at (2.2555,0.7600) {185 correct};
\path[seg, fill=wrong] (3.0111,0.5600) rectangle (4.9469,0.9600);
\node[lab, text=white] at (3.9790,0.7600) {237 wrong};
\path[recov] (4.9469,0.5600) rectangle (6.3600,0.9600);
\node[lab, text=ink, fill=white, rounded corners=0.5pt] at (5.6535,0.7600) {173};
\node[anchor=north west, inner sep=0pt, align=left, font=\fsS] at (6.4600,0.9700) {31.1\% correct\\[-1pt](Selective: 31.9\%)};
\path (0,0.30) -- (0,3.84);

\end{tikzpicture}
\endgroup
\par\vspace{4pt}
\begingroup
\definecolor{flowWrong}{HTML}{4B5563}
\newcommand{\legendfont}{\footnotesize}
\begin{tikzpicture}[inner sep=0pt,outer sep=0pt]
  \fill[gcGt] (0,-0.07) rectangle (0.18,0.07);
  \fill[gcRt] (0.18,-0.07) rectangle (0.36,0.07);
  \fill[gcFt] (0.36,-0.07) rectangle (0.54,0.07);
  \node[font=\legendfont,anchor=west,xshift=3pt] (correct) at (0.54,0) {Correct (by stage)};
  \fill[flowWrong] ([xshift=7pt]correct.east) ++(0,-0.07) rectangle ++(0.18,0.14);
  \node[font=\legendfont,anchor=west] (wrong) at ([xshift={10pt+0.18cm}]correct.east) {Wrong};
  \path[draw=flowWrong,fill=white,line width=0.35pt,
        postaction={pattern=north east lines,pattern color=flowWrong!55}]
    ([xshift=7pt]wrong.east) ++(0,-0.07) rectangle ++(0.18,0.14);
  \node[font=\legendfont,anchor=west] at ([xshift={10pt+0.18cm}]wrong.east)
    {Wrong after prior correct};
\end{tikzpicture}
\endgroup
\caption{\textbf{Request flow on OSG-R (left) and UIV (right)}, Q36/Q38/Q38, drawn as the SSP flow figure of the main paper. Regrounder exits combine Detail and Context agreement (\V{mixosg}{ctxReq} and \V{mixuiv}{ctxReq} Context exits). On UIV the Finalizer leaves \V{mixuiv}{fAfter}\% of its requests correct, against \V{mixuiv}{fBefore}\% for Selective.}
\label{fig:s_flow}
\end{figure*}
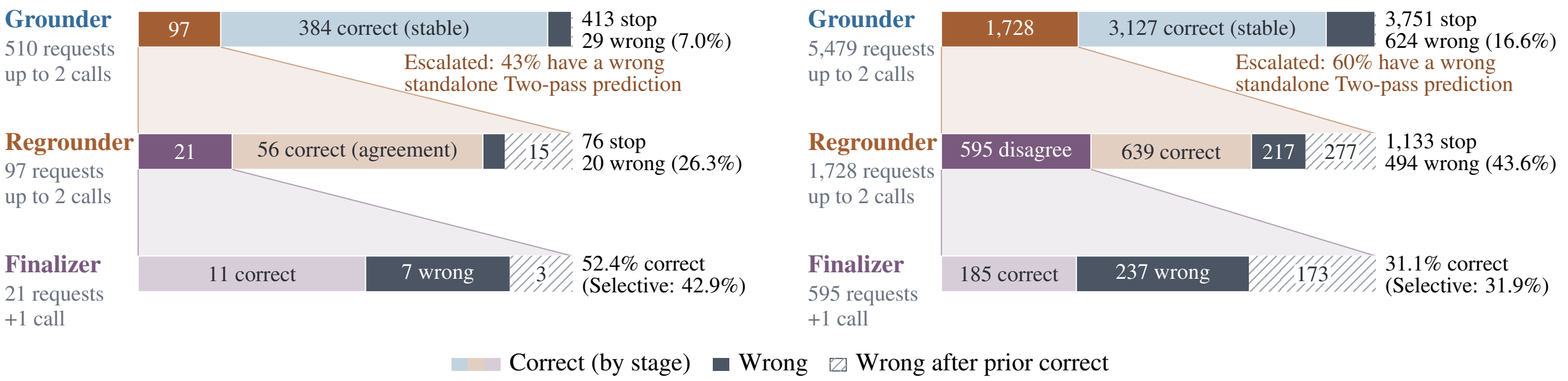

Figure~\ref{fig:s_flow} traces requests through the cascade on OSG-R and UIV, extending the SSP diagnostic in the main paper.

\end{document}